\documentclass[acmtog]{acmart}

\usepackage{booktabs} 
\usepackage{pifont}
\newcommand{\cmark}{\ding{51}}
\newcommand{\xmark}{\ding{55}} 
\usepackage{float}
\usepackage{multirow}
\usepackage{appendix}
\usepackage{xcolor}

\usepackage[ruled]{algorithm2e} 

\SetAlFnt{\small}
\SetAlCapFnt{\small}
\SetAlCapNameFnt{\small}
\SetAlCapHSkip{0pt}

\acmJournal{TOG}

\newcommand{\myparagraph}[1]{\vspace{0.75ex}\par\noindent \textit{\textbf{#1}}}
\newcommand{\afterfigure}{\vspace{-1em}}
\newcommand{\aftertable}{\vspace{-2em}}

\copyrightyear{2025}
\acmYear{2025}
\makeatletter
\def\@ACM@copyright@check@cc{}
\makeatother
\setcopyright{cc}
\setcctype{by}
\acmConference[SA Conference Papers '25]{SIGGRAPH Asia 2025 Conference Papers}{December 15--18, 2025}{Hong Kong, Hong Kong}
\acmBooktitle{SIGGRAPH Asia 2025 Conference Papers (SA Conference Papers '25), December 15--18, 2025, Hong Kong, Hong Kong}\acmDOI{10.1145/3757377.3763838}
\acmISBN{979-8-4007-2137-3/2025/12}
\newcommand{\notation}[1]{\ensuremath{#1}\xspace}

\newcommand{\InputImages}{\notation{\mathbf{I}}}
\newcommand{\InputImage}{\notation{I}}
\newcommand{\TargetImages}{\notation{\mathbf{T}}}
\newcommand{\TargetImage}{\notation{T}}
\newcommand{\RenderedImages}{\notation{\mathbf{\hat{T}}}}
\newcommand{\Real}{\notation{\mathbb{R}}}
\newcommand{\Rays}{\notation{\mathbf{R}}}
\newcommand{\Ray}{\notation{R}}
\newcommand{\PatchSize}{\notation{p}}

\newcommand{\Tokens}{\notation{\mathbf{M}}}
\newcommand{\Token}{\notation{M}}
\newcommand{\TokenIndex}{\notation{m}}
\newcommand{\Conv}{\notation{\text{Conv}}}
\newcommand{\Concat}{\notation{\text{Concat}}}
\newcommand{\ChannelDim}{\notation{d}}
\newcommand{\Neighbors}{\notation{\mathcal{N}}}
\newcommand{\Conditioning}{\notation{{C}}}
\newcommand{\LayerIndex}{\notation{l}}
\newcommand{\Key}{\notation{\mathbf{K}}}
\newcommand{\Query}{\notation{\mathbf{Q}}}
\newcommand{\Value}{\notation{\mathbf{V}}}
\newcommand{\Weight}{\notation{W}}
\newcommand{\WeightQ}{\notation{\Weight_q}}
\newcommand{\WeightK}{\notation{\Weight_k}}
\newcommand{\WeightV}{\notation{\Weight_v}}
\newcommand{\Window}{\notation{w}}
\newcommand{\Opacity}{\notation{\sigma}}
\newcommand{\ViewDir}{\notation{\hat{\mathbf{n}}}}

\begin{document}
\title{LVT: Large-Scale Scene Reconstruction via Local View Transformers}

\author{Tooba Imtiaz$^{1,2}$}
\authornote{denotes equal contribution}
\email{imtiaz.t@northeastern.edu}
\author{Lucy Chai$^{1}$}
\authornotemark[1]
\email{lucyrchai@google.com}
\author{Kathryn Heal$^{1}$}
\email{kheal@google.com}
\author{Xuan Luo$^{1}$}
\email{xuluo@google.com}
\author{Jungyeon Park$^{1}$}
\email{jungyeonp@google.com}
\author{Jennifer Dy$^{2}$}
\email{jdy@ece.neu.edu}
\author{John Flynn$^{1}$}
\email{jflynn@google.com.}
\affiliation{ $^1$Google, USA, $^2$Northeastern University\country{USA}}

\begin{abstract}

Large transformer models are proving to be a powerful tool for 3D vision and novel view synthesis. However, the standard Transformer's well-known quadratic complexity makes it difficult to scale these methods to large scenes. To address this challenge, we propose the Local View Transformer (LVT), a large-scale scene reconstruction and novel view synthesis architecture that circumvents the need for the quadratic attention operation. Motivated by the insight that spatially nearby views provide more useful signal about the local scene composition than distant views, our model processes all information in a local neighborhood around each view. To attend to tokens in nearby views, we leverage a novel positional encoding that conditions on the relative geometric transformation between the query and nearby views. We decode the output of our model into a 3D Gaussian Splat scene representation that includes both color and opacity view-dependence.
Taken together, the Local View Transformer enables reconstruction of arbitrarily large, high-resolution scenes in a single forward pass. See our project page for results and interactive demos: \href{https://toobaimt.github.io/lvt/}{https://toobaimt.github.io/lvt/}.
\end{abstract}

%
%

\begin{teaserfigure}
  \includegraphics[width=0.95\textwidth]{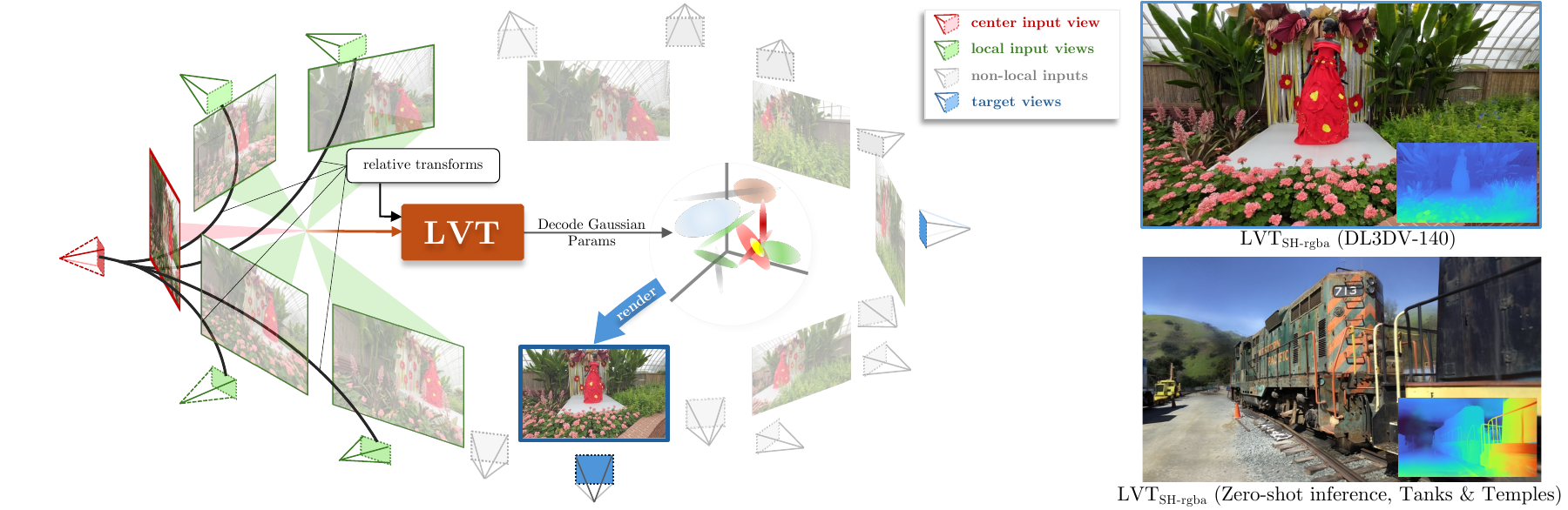}
  \caption{\textbf{Local View Transformer (LVT) for scalable scene reconstruction.} Instead of the quadratic self-attention operation typically used in transformers, LVT processes information within local neighborhoods relative to each view, based on the insight that spatially nearby views are generally more informative about the local scene structure than distant ones. Cascading these LVT blocks increases the effective receptive field. Our model takes a set of input views representing the entire scene and, in a single feed-forward pass, outputs a Gaussian splat scene representation which can then be rendered to various target cameras. LVT scales linearly, rather than quadratically, with the number of input images, and generalizes to varied sequence lengths and out-of-distribution camera trajectories.}
  \vspace{.1in}
  \label{fig:teaser}
\end{teaserfigure}

\maketitle

\section{Introduction}



Echoing their widespread impact across other domains~\cite{devlin-etal-2019-bert, radford2019language, caron2021emerging,carion2020end,ranftl2021vision,dosovitskiy2020vit}, the recent application of Transformer~\cite{vaswani2017attention}-based architectures to novel view synthesis (NVS) has achieved remarkable success~\cite{hong2023lrm,zhang2024gs,ziwen2024llrm}. Unlike earlier techniques that relied more heavily on explicit 3D reasoning, these models instead opt for a primarily data-driven approach to infer scene structure from input images.
Despite their state-of-the-art performance, a significant drawback of transformers remains their high computational cost, primarily driven by the self-attention mechanism's quadratic scaling with token count. Even with optimizations such as Flash Attention~\cite{dao2022flashattention}, global self-attention remains costly for long sequences. This computational burden poses a significant bottleneck to applying these models to large real-world scenes with many high resolution input views.

In contrast, per-scene optimization approaches, like 3DGS~\cite{kerbl20233d}, are able to perform scene-scale reconstruction but require thousands of test-time gradient steps to do so.
Other methods ~\cite{flynn2024quark,flynn2016deepstereo} sidestep a persistent global scene representation and opt instead to reconstruct local geometry for each rendered view on-the-fly as the camera moves through the scene. Although this strategy is efficient, the lack of a global model results in noticeable flickering as the camera moves. 

Recently, Long-LRM ~\cite{ziwen2024llrm} addresses the computational cost of transformers by employing a hybrid architecture that integrates Mamba2 \cite{mamba2} and transformer blocks. The method allows scaling to a considerably larger number of input views than prior transformer-based scene-reconstruction methods. However, the approach requires several optimizations for efficiency, such as heuristic Gaussian merging and pruning. It also lacks natural generalization to varied-length sequences, and needs camera selection to maintain a fixed number of inputs. This highlights the need for more scalable and robust generalizable reconstruction techniques.

To this end, we propose the Local View Transformer (LVT), an architecture designed for efficient and high-fidelity 3D Gaussian splat reconstruction at the scene scale in a single inference pass. In contrast with prior global attention-based transformer models for 3D reconstruction, LVT exploits the inherent spatial locality in 3D scenes. We motivate our design by drawing a parallel to Convolutional Neural Networks~\cite{lecun2015deep} (CNNs). CNNs largely replaced earlier neural network architectures based on dense connections with a local sliding window mechanism. The CNN layer brought several key advantages: enhanced efficiency, improved generalization via translation invariance, and the ability to scale to arbitrary-sized inputs. While an individual CNN's receptive field is limited, stacking multiple CNN layers progressively widens the receptive field, enabling broader contextual understanding.

In LVT, we extend this localized-processing paradigm to the attention mechanism within a transformer. Specifically, our attention mechanism operates locally, limiting tokens from an input or query view to attend \emph{only} to tokens within a small set of neighboring views. The definition of "neighboring" can vary; for a scene capture containing sequential images, this might involve attending to the previous and next views. For a more general collection of views, this could involve choosing views based on the distance between camera centers. Although each LVT transformer block limits attention to nearby views, stacking multiple blocks progressively expands the `view receptive field,' in a similar manner to a stack of CNN layers.

Mirroring the translation invariance inherent in CNNs, our method is designed to be \emph{transformation} invariant to the 3D coordinate frame of the query view. This means that tokens within each view encode splat geometry relative to that view's local coordinate frame. Consequently, when attending from a query view to a neighboring view, the token features in the neighboring view must be transformed into the query view's coordinate frame. To achieve this we incorporate a straightforward encoding of the relative poses between views, and found this sufficient to achieve the desired invariance. This invariance allows LVT to transfer priors learned from shorter sequences during training to arbitrarily long sequences during inference.  


Finally, we introduce view-dependent opacity, allowing for a richer representation of specular highlights, dynamic lighting effects, and thin objects within scene reconstructions, in contrast to prior methods that solely encode view-dependent colors~\cite{kerbl20233d,mildenhall2021nerf,fridovich2022plenoxels}. 

Altogether, our LVT model surpasses prior methods to achieve state-of-the-art novel view synthesis on large scenes. 
Quantitatively, on the challenging DL3DV benchmark \cite{ling2024dl3dv}, LVT outperforms the previous state-of-the-art, Long-LRM \cite{ziwen2024llrm}, by a significant margin of {+3.54dB} PSNR and 3DGS, which is optimized per scene, by {+2.11dB} PSNR. Our model also exhibits zero-shot generalization capabilities on the Tanks\&Temples dataset \cite{knapitsch2017tanks}, outperforming Long-LRM by {+2.79dB PSNR, and per-scene optimized 3DGS by +0.72dB PSNR}. In summary:
\begin{itemize}
    \item We propose LVT, a transformer architecture for feed-forward novel view synthesis of large scenes that scales linearly to long and varied-length sequences.
    \item Our model introduces a local attention module in which image tokens attend only to tokens from nearby views thus eliminating the costly self-attention operation. To achieve this local processing we condition the tokens on relative camera poses, 
    \item  For further improvements, we incorporate view-dependent opacity prediction to enhance rendering fidelity, particularly in scenes with thin structures and complex lighting effects.
    \item Our extensive experiments and ablations demonstrate an effective, scalable approach for novel view synthesis of full scenes in a single inference pass. 
\end{itemize}

\begin{figure*}
  \includegraphics[width=0.9\linewidth]{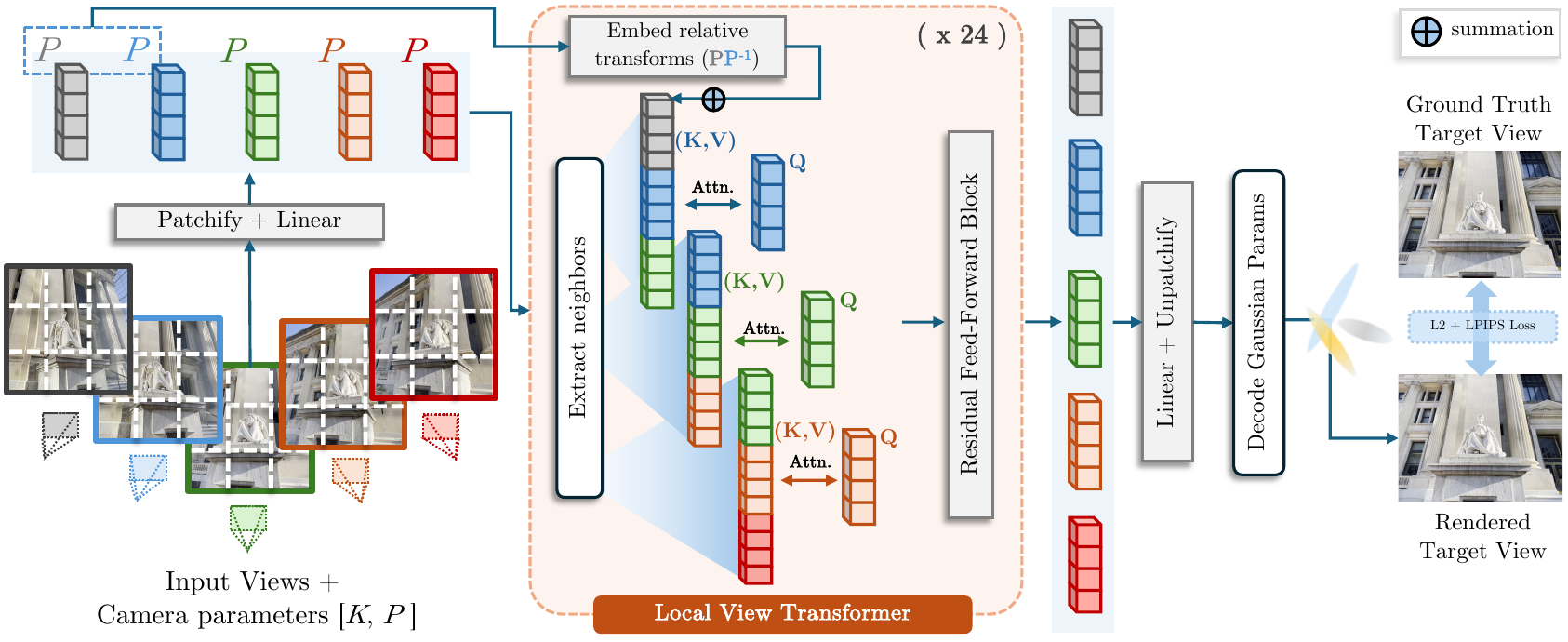}
  \caption{\textbf{Overview of Local View Transformer architecture.} The pipeline takes input view sequences and the corresponding local ray maps, and patchifies them to obtain input tokens. Within each LVT block, the tokens from a window of $\Window$ neighbor views (here $\Window=3$ for illustration) relative to a query view are consolidated and updated with the corresponding relative transformation embeddings. The tokens from the \emph{query} view then selectively attend to these neighbor tokens (instead of attending to all input tokens, as full self-attention does). This LVT block is repeated 24 times. The processed tokens are then unpatchified and decoded to pixel-aligned Gaussian splat parameters. The combined Gaussian splats are rendered using the 3DGS renderer and compared to ground truth during training.}
  \label{fig:arch-main}
  \afterfigure
\end{figure*}
\section{Related Work}

3D reconstruction and novel view synthesis have long been central challenges in computer vision. Early works used custom algorithms and heuristics to generate a scene representation from a set of input images ~\cite{kutulakos2000theory,seitz1999photoconsistency,agarwal2009building,chaurasia2013depth, penner2017soft}.  
In contrast, ~\cite{ mildenhall2021nerf, kerbl20233d} leverage differentiable rendering to directly optimize a dedicated 3D representation for individual scenes. These techniques deliver high fidelity at the expense of requiring substantial test-time optimization, typically ranging from minutes to hours. Additionally, since these methods optimize on a per-scene basis, they do not learn generalizable priors applicable across different scenes. Consequently, to prevent over-fitting, they require dense input views or the addition of ad-hoc regularization losses.

\myparagraph{Feed-forward models.}
Generalizable methods avoid these drawbacks by predicting the entire scene in a single feed‑forward pass though a deep learning pipeline ~\cite{flynn2016deepstereo,yifan2019differentiable,aliev2020neural,lin2020deep,zhou2018stereo}.  
One approach for these methods is to leverage explicit 3D geometric structures, such as epipolar lines \cite{suhail2022generalizable, wang2021ibrnet} or plane-sweep cost volumes \cite{flynn2016deepstereo, mildenhall2019llff, flynn2019deepview,  liu2020neural, chen2021mvsnerf, johari2022geonerf, chen2024mvsplat, xu2024depthsplat, flynn2024quark}, to aggregate information from input views. While these methods can be effective for a small number of inputs views from forward-facing cameras, they lack a unified representation and and typically resort to transitioning between local representations to process larger, general scenes.

\myparagraph{Large reconstruction models.} 
With the rise of transformers, view synthesis methods have moved away from explicit 3D inductive biases towards fully learned, data-driven geometric priors, culminating in Large Reconstruction Models (LRMs). 
Existing LRMs employ two primary output strategies: either decoding to an explicit 3D representation, such as triplane NeRF fields~\cite{hong2023lrm,li2023instant3d,wang2023pf} or pixel‑aligned 3‑D Gaussian splats~\cite{zhang2024gs,tang2024lgm}, or directly decoding to target pixels, as exemplified by LVSM~\cite{jin2025lvsm}.
These models achieve impressive results from extremely sparse inputs—often one to four images with little or no view overlap—by leveraging the global context captured by transformers.  However, the quadratic complexity of transformer attention poses a significant practical challenge to scaling these models to real-world scenes with numerous input views or high-resolution imagery. Thus, these prior models focus on short, low-resolution captures.

\myparagraph{Scaling to longer input sequences.}
Recent work, such as Long‑ LRM \cite{ziwen2024llrm} and Fast3R \cite{yang2025fast3r}, explore scalable NVS architectures that can handle longer input sequences.
Long‑LRM~\cite{ziwen2024llrm} mitigates the computational cost of transformers by combining Mamba2~\cite{pmlr-v235-dao24a} and transformer blocks, pushing prior feed‑forward LRMs~\cite{zhang2024gs,jin2025lvsm} from roughly 2–4 to 32 input views. However, the use of Mamba2 sacrifices the standard transformer model's natural order invariance. This property is particularly beneficial for view synthesis, as input views often lack a definitive ordering. Fast3R~\cite{yang2025fast3r} generalizes DUST3R~\cite{wang2024dust3r} from two input views to longer sequences for camera registration. However, both models at their core still adopt full self-attention, which becomes computationally infeasible at for a large number of input views. Spann3R~\cite{wang20243dreconstructionspatialmemory} and CUT3R~\cite{wang2025continuous3dperceptionmodel} build the scene incrementally using a memory bank or a recurrent state.

In contrast, our proposed Local View Transformer explores an alternative approach to handling long and variable-length input sequences, inspired by the spatial nature of 3D scenes. Unlike prior methods, LVT replaces \emph{global} full self-attention with a more \emph{localized} attention operation across tokens from neighboring views. 
We find that this strategy largely maintains the benefits of the attention module while scaling linearly with the number of input views.


\myparagraph{Positional encoding.} 
Transformers are by design permutation invariant, so positional encodings are used to provide the model with information about the order of input tokens.
Approaches such as RoPE have explored using relative, rather than absolute, positional embeddings to improve scalability and generalization~\cite{heo2024rotary,su2024roformer}. 
Analogously, recent works have explored embedding camera information as a relative positional encoding for geometric tasks to allow for invariance towards the choice of a global reference frame~\cite{kong2024eschernet,li2025cameras,miyato2023gta}. Our Local View Transformer also adopts relative camera poses as a positional embedding signal; combined with our windowed attention mechanism, this maintains transformation invariance in the local neighborhoods for a more scalable transformer architecture over long input sequences. 


\myparagraph{View dependence in 3D reconstruction.} Accurately modeling view-dependent effects, such as specular reflections and highlights, has been a consistent challenge in novel view synthesis. Ref‑NeRF~\cite{verbin2022ref} improves upon NeRF's ~\cite{mildenhall2021nerf} basic view dependence by factorizing color into diffuse and specular components, yielding more accurate reflections particularly those with complex lighting and reflective materials. Drawing from their application in graphics, various approaches~\cite{wizadwongsa2021nex,fridovich2022plenoxels,yu2021plenoctrees} employ spherical harmonics~\cite{ramamoorthi_sh} to model view-dependent effects. Spherical harmonics simplify the calculation of view-dependent color to a straightforward dot product between the viewing direction and the predicted coefficients.


While the use of view dependent color to model directional effects is widespread, the use of view-dependent opacity has been relatively limited in prior work, possibly due to overfitting risks for methods that employ per-scene optimization. Gaussian Opacity Fields (GOF) \cite{yu2024gaussian} perform  heuristic merging of per-view opacity. VoD-3DGS \cite{nowak2025vod}, uses a learnable per-Gaussian symmetric matrix that modulates opacity based on viewing direction. However, their view-dependent appearance model struggles to reconstruct scenes with low specular content, and suffers from degraded quantitative performance.

In our approach, we also include view-dependent opacity, finding that a straightforward application of spherical harmonics to opacity substantially improves rendered quality, especially on long input sequences.

\section{Method}\label{sec:method}
The Local View Transformer adapts the transformer architecture to process tokens in a locally-aware fashion. Below, we detail our model components and their differences from the standard Transformer. 



\subsection{Architecture Overview}\label{sec:method_architecture}
As illustrated in Figure \ref{fig:arch-main}, given $N_i$ input images \InputImages with known camera intrinsics and poses, i.e. $\InputImages=\{\InputImage_j \in \Real^{H\times W\times 3}, K_j, P_j\}, j\in\{1,...,N_i\}$, the Local View Transformer (LVT) predicts the parameters of input pixel-aligned 3D Gaussian splats. We train the network end-to-end by rendering from these splats to $N_t$ target views $\TargetImages=\{\TargetImage_k \in \Real^{H\times W\times 3}, K_k, P_k\}, k \in \{1,...,N_t\} $ and comparing the rendered target images to the corresponding ground truth images. Our architecture mimics the structure of GS-LRM~\cite{zhang2024gs}, beginning with tokenization into image patches, followed by several transformer blocks, and ending with a decoding block that converts the tokens into per-pixel Gaussian Splat parameters~\cite{kerbl20233d}. However, we make modifications to ensure \emph{transformation invariance} at each step, i.e. all processing is performed locally, relative to each input view, and independent of an arbitrary world coordinate system. 

\subsection{Tokenizing Images to Patches} Following prior works~\cite{dosovitskiy2020vit}, we first patchify the inputs into a series of non-overlapping tokens with patch-size $\PatchSize$. The conventional practice in related LRM models~\cite{zhang2024gs, jin2025lvsm, ziwen2024llrm} concatenates input pixels with Pl\"ucker ray coordinates \cite{plucker1865xvii} to encode input camera parameters along the channel dimension, which are converted to tokens representing each $\PatchSize \times \PatchSize$ patch via a convolutional kernel. 

In contrast, as our LVT model processes views \textit{locally}, our input tokens do not need to capture where the view is situated in 3D space.  Rather than Pl\"ucker rays which encode both the camera intrinsics and extrinsics, we compute the set of normalized local ray maps $\Rays = \{\Ray_j \in \mathbb{R}^{H\times W\times3}, j = 1, ..., N_i\}$ using just the intrinsics $K_j$.
These local ray maps are concatenated to the image pixels and patchified into a set of tokens $\Tokens \in \Real^{N_i \times \frac{H}{\PatchSize} \times \frac{W}{\PatchSize} \times \ChannelDim}$ using a convolutional kernel:
\begin{equation}\label{eqn:patchify}
    \begin{aligned}
        \Token_j &= \Conv ( \Concat ([\InputImage_j,\Ray_j])) \\
        \Tokens &= \{\Token_j | j \in 1, ..., N_i\}, \\
    \end{aligned}
\end{equation}
where \ChannelDim refers to the channel dimension. These tokens are then flattened to shape $N_i \times \frac{HW}{\PatchSize^2} \times \ChannelDim$ and normalized prior to the transformer blocks.

\subsection{Local View Transformer}

The core idea of our Local View Transformer is that nearby image neighbors provide the most relevant information for a given view. Thus, rather than computing the full quadratic self-attention matrix in the standard Transformer architecture, we only need to compute attention between a query view and its neighbors.

In more detail, for a given image $\InputImage_j$ associated with tokens $\Token_j$ with shape ${\frac{HW}{\PatchSize^2} \times \ChannelDim}$, we first define the set of neighbors $\Neighbors(j)$, where each image $j$ is also in its own neighbor set. To provide a transformation invariant spatial conditioning signal, instead of directly using camera poses $P_j$, we use the relative poses $P_{j} P_{j'}^{-1}$ between the $j$-th and $j'$-th input views. We then apply a positional encoding~\cite{vaswani2017attention} on the relative quaternions and translations.
The positionally-encoded values are next processed by several residual MLP blocks to produce a set of conditioning features $\Conditioning_{(j, j')} \in \Real^\ChannelDim$ that capture the relative spatial information between neighboring views. Relatedly, rotary positional embeddings~\cite{heo2024rotary,su2024roformer} compute transformations on the key and query feature spaces to encode the relative rather than absolute positions of transformer tokens, whereas our conditioning features operate on the relative rigid-body transformation between neighboring camera poses. Several heuristics, including nearest $\Window$ spatial neighbors, can be used to determine the set of neighbors $\Neighbors(j)$. {In our experiments, we use the distance between camera centers to select the neighbors. We discuss alternative choices in our ablation study.}

Next, we modify the transformer attention block to attend only to neighboring tokens and condition on the relative transformation feature. We use the notation $\Token_j^{(\TokenIndex, \LayerIndex)}$ to denote the $\TokenIndex$-th token at the $\LayerIndex$-th layer corresponding to input image $j$, where $\TokenIndex \in 1, ..., \frac{HW}{\PatchSize^2}$, and  $\Conditioning_{(j, j')}^{(\LayerIndex)}$ as the conditioning feature at that layer, transformed via a per-layer Dense projection. We compute: 
\begin{equation}
\begin{aligned}
\Query_j^{(\TokenIndex, \LayerIndex)} &= \WeightQ  \Token_j^{(\TokenIndex, \LayerIndex)} \\
\Key_{j'}^{(\TokenIndex', \LayerIndex)} &= \WeightK \left(\Token_{j'}^{(\TokenIndex', \LayerIndex)} + \Conditioning_{(j, j')}^{(\LayerIndex)}\right) \\
\Value_{j'}^{(\TokenIndex', \LayerIndex)} &= \WeightV \left(\Token_{j'}^{(\TokenIndex', \LayerIndex)} + \Conditioning_{(j, j')}^{(\LayerIndex)}\right)  \\
\end{aligned}
\end{equation}
In essence, the keys and values contain not only information about the scene structure from the neighbors in $\Token_{j'}^{(\TokenIndex', \LayerIndex)}$ but also information about the relative positioning between neighbors in $\Conditioning_{(j, j')}$. The resulting feature becomes:
\begin{equation}
\Token_j^{(\TokenIndex,\;\LayerIndex+1)} = \underset{j' \in \Neighbors(j),\;{m}' \in 1,...,\frac{HW}{\PatchSize^2}}{\text{softmax}}\left(\frac{\Query_j^{(\TokenIndex, \LayerIndex)}\Key_{j'}^{(\TokenIndex', \LayerIndex)T}}{\sqrt{\ChannelDim}}\right)\Value_{j'}^{(\TokenIndex', \LayerIndex)}
\end{equation}

The remainder of LVT follows the Vision Transformer architecture, interleaving our modified residual neighborhood attention blocks with standard residual feed-forward blocks. We use pre-layer normalization ~\cite{on_layer_normalization} before each attention and feed-forward layer and use Flash Attention~\cite{dao2022flashattention} within the attention blocks. We note that cascading the LVT blocks effectively grows the receptive field of the tokens, expanding the scope of attention to more views with each subsequent layer.

Following the transformer blocks, each processed token, $\Token_j^{(\TokenIndex,\;L)}$ at the final layer $L$ is unpatchified using a transposed convolution layer into a patch size of $\PatchSize \times \PatchSize$ with $\ChannelDim_G$ channels to yield pixel-aligned Gaussian parameters, where $\ChannelDim_G$ is the total number of Gaussian parameters. At each pixel, we parameterize the corresponding 3D Gaussian by a 3-channel scale, a 4-channel  quaternion, scalar depth along the pixel's ray, $\ChannelDim_c$-channel color and $\ChannelDim_o$-dimensional opacity $\Opacity$. The translation, rotation and scale are relative to the corresponding view's 3D coordinate frame and are transformed into the world coordinate frame for rendering. During training, the decoded $N_i\times H \times W$ Gaussians are rendered to target views using the differentiable 3DGS rasterizer \cite{kerbl20233d}. 

\subsection{View-Dependent Splat Opacity}\label{sec:view-dep-op}

Previous works have used spherical harmonics to represent view-dependent effects in both 3D Gaussian Splat and other neural representations \cite{sloan2023precomputed, fridovich2022plenoxels, muller2022instant}. We extend the application of spherical harmonics coefficients from color representation to also compute view-dependent opacity. 
{Although view-dependent opacity has been less frequently utilized in prior work~\cite{nowak2025vod,heitz2015sggx}, potentially due to over-parameterization concerns, we note that LVT inherits the generalizability of feed-forward methods without overfitting.
We find that this addition of view-dependent opacity better captures thin structures and sharp boundaries especially over long input sequences. This allows for a representation that works well locally and transitions smoothly as the camera changes.
}
As a result, each view-dependent splat in our model is parameterized by $\ChannelDim_c = 3\times(deg\_SH_{color}+1)^2$ and $\ChannelDim_o = (deg\_SH_{\Opacity}+1)^2$ channels for color and opacity respectively. The spherical harmonics are computed relative to each view's coordinate frame, {consistent with our focus on transformation invariance,} and are transformed to world space for rendering.

\myparagraph{Regularization.} To avoid spurious opacities when viewed from novel viewing angles, 
we introduce a regularization term that penalizes extreme magnitudes of splat opacity evaluated at randomly sampled viewing directions on the surface of a unit sphere. Let $\ViewDir$ be a randomly sampled direction encoded with the spherical harmonic basis. We define the regularizer
\begin{equation}\label{eqn:alpha_reg}
 \begin{aligned}
    R_\sigma = \frac{1}{HWN_i}\sum_{n=1}^{HWN_i}|\Opacity_{(n)} \cdot \ViewDir | 
 \end{aligned}
\end{equation}
where $\Opacity_{(n)}$ refers to the opacity coefficients corresponding to the $n$-th pixel. In our models that do not use view-dependent opacity, we simply regularize $R_\sigma = \frac{1}{HWN_i}\sum_{n=1}^{HWN_i}|\Opacity_{(n)}|$ following Long-LRM~\cite{ziwen2024llrm} where $\Opacity_{(n)}$ is a per-pixel scalar opacity. 

\subsection{Training Objective}\label{sec:method_training}
We render the predicted Gaussian splats using the target cameras and minimize the loss between the rendered images  $\RenderedImages$ and the ground truth target views $\TargetImages$. Our reconstruction loss function $\mathcal{L}$ is a weighted combination of the Mean Squared Error (MSE) loss, Perceptual loss, and opacity regularization: 

\begin{align}\label{eqn:loss}
    \mathcal{L} = \frac{1}{N_t} \sum_{k=1}^{N_t}\left(\mathrm{MSE}\left(\hat{T}_k,T_k\right) + \lambda \cdot \mathrm{Perceptual}\left(\hat{T}_k,T_k\right) \right) + \alpha \cdot R_\sigma,
\end{align}
where $\lambda$ is the weighting factor for the perceptual loss and $\alpha$ is a weighting factor for the opacity regularizer. Following prior LRMs \cite{hong2023lrm, xu2023dmv3d, wang2023pf}, we adopt LPIPS \cite{zhang2018unreasonable} as the perceptual loss, in contrast to ~\cite{zhang2024gs} which uses a VGG-based~\cite{simonyan2014very} loss for training stability. 
\section{Experiments}

We provide training and implementation details of our method (Sec.~\ref{sec:experiments-implementation}), comparison against baselines (Sec.~\ref{sec:experiments-baselines}), and ablate the effect of various design choices (Sec.~\ref{sec:experiments-ablations}). We include additional experiments, including a model trained for 2K resolution, more ablations, and video results in our supplementary material.

\begin{table*}[t!]
    \centering
    \begin{tabular}{l c |c c c|c c c | c c c} 
         \multirow{2}{*}{Model} &  
         \multirow{2}{*}{\shortstack[c]{Feed-\\Forward}} & 
         \multicolumn{3}{c|}{DL3DV-140} &
        \multicolumn{3}{c|}{Tanks\&Temples} & 
        \multicolumn{3}{c}{MipNeRF360} \\
          & & \textit{PSNR} $\uparrow$ & \textit{SSIM}  $\uparrow$ & \textit{LPIPS}  $\downarrow$ & \textit{PSNR} $\uparrow$ & \textit{SSIM}  $\uparrow$ & \textit{LPIPS}  $\downarrow$ &
         \textit{PSNR} $\uparrow$ & \textit{SSIM}  $\uparrow$ & \textit{LPIPS}  $\downarrow$\\
         \midrule
         3DGS$_{30k}$ & \xmark           & \underline{25.53} & \underline{0.849} & \underline{0.210} & \underline{20.45} & \underline{0.780} & \underline{0.251} & \textbf{24.81} & \textbf{0.745} & \textbf{0.246} \\
         Long-LRM  & \cmark      & {24.10} & {0.783} & {0.254} & {18.38} & {0.601} & {0.363}  & -- & -- & -- \\
         \midrule
         LVT$_{base}$  & \cmark &  {23.65} & {0.813} & {0.215} & {18.33} & {0.681} & {0.296} & {21.51} & {0.603} & {0.335} \\
         LVT$_{SH-rgba}$ & \cmark  & {\textbf{27.64}} & {\textbf{0.883}} & {\textbf{0.133}} & {\textbf{21.17}} & {\textbf{0.794}} & {\textbf{0.174}} & {\underline{24.53}} & {\underline{0.698}} & {\underline{0.251}} \\
    \bottomrule
    \end{tabular}
    \caption{\textbf{Comparison of LVT against baselines.} We evaluate at $960\times540$ resolution on the DL3DV-140 test set and perform zero-shot evaluation on Tanks\&Temples and MipNeRF360.
     Our LVT$_{SH-rgba}$ model outperforms both baselines across all metrics for DL3DV140, and perform competitively with 3DGS in a single feed-forward pass on the latter benchmark datasets. Our model is able to generalize to Tanks\&Temples and MipNeRF360 datasets, despite differences in the sequence lengths and camera trajectories than those used for training. We obtain numbers of Long-LRM from their paper as the model is not publicly available, and note that Long-LRM is unable to generalize to Mip-NeRF360. We retrain 3DGS for each scene using the same input views as our model.
     }\label{tab:results_main}
     \aftertable
\end{table*}

\subsection{Implementation Details}\label{sec:experiments-implementation}

\myparagraph{Datasets}. We train our model on DL3DV~\cite{ling2024dl3dv}, a large-scale dataset of 4K resolution videos captured from bounded and unbounded real-world scenes for benchmarking novel view synthesis. The training and test splits, DL3DV-10K and DL3DV-140, consist of 10,510 and 140 videos, respectively. The camera poses in each video are annotated using COLMAP~\cite{schonberger2016pixelwise}. For training, we sample a sequence of input images separated by a random stride between four and eight, and sample target images between the inputs. {We construct the neighborhood graph by connecting each image $\InputImage(j)$ to its closest neighbors $\Neighbors(j)$ based on camera center distances.} 

We evaluate on the DL3DV test split and additionally perform zero-shot evaluation on Tanks\&Temples~\cite{knapitsch2017tanks} (the train and truck scenes) and  Mip-NeRF360~\cite{barron2022mipnerf360}. We use every eighth image in the sequence as targets. On DL3DV, we take every eighth image of the remaining images after removing the targets; as such, the number of input images varies with the length of the image sequence, and the distance of the targets to the nearest input also varies. On Tanks\&Temples and Mip-NeRF360, we reduce our input stride to every fourth image after removing targets due to the larger camera spacing. We use temporal order for neighborhood graph on DL3DV and Tanks\&Temples; for Mip-NeRF360 we compute neighbors based on camera center distances, as the cameras capture an upper hemisphere rather than a video flight trajectory. In the supplementary (Sec.~\ref{sec:supp-re10k}), we also train and evaluate a model on RealEstate10k~\cite{zhou2018stereo}.





\myparagraph{Training details.} We follow the basic transformer structure of GS-LRM~\cite{zhang2024gs}, using 24 transformer layers, with hidden dimensions of 1024 for the transformer and 4096 for the MLP. We use 16 heads for the attention layer.  We scale the initialization of the final kernels of each residual block (in the attention and MLP modules) by $\sqrt{\frac{1}{N_r}}$ where ${N_r}$ is the total number of residual connections, and otherwise use LeCun Normal initialization on the learnable parameters and omit biases. We use a peak learning rate of 2e-4 for 150K steps, consisting of 2K steps linear warmup on the learning rate and cosine decay for the remaining steps. We first train at a maximum resolution of $480\times270$, and then finetune at $960\times540$ for an additional 20K steps with peak learning rate 1e-5 following 2K steps of linear warmup. We use loss weights $\lambda=0.05$ and $\alpha=0.001$, and batch size 256.


\myparagraph{Multi-resolution, multi-length dataset mixing.} As the computational costs of the transformer and rendering scale with the resolution and number of input images, we use a dataset mixing strategy to allow the model to generalize to longer sequences of varied length. That is, we train on three resolutions of the dataset: $240\times135$, $480\times270$, and $960\times540$. We train on the longest sequences at the low $240\times135$ resolution, sampling between 40 and 52 input images. At the middle $480\times270$ resolution, we sample up to 24 input images, and at $960\times540$ resolution we sample 16 input images. Additional details are provided in the supplementary (Tab.~\ref{tab:supp_dataset}). 

\subsection{Comparison Against Baselines}\label{sec:experiments-baselines}

We train several versions of our LVT model with variations in the spherical harmonics coefficients. Our base model (LVT$_{base}$) does not have any view-dependence in the representation, while our full model (LVT$_{SH-rgba}$) uses view-dependence on both color and opacity. We compare our results to Long-LRM~\cite{ziwen2024llrm}, which uses full self-attention and thus does not naturally generalize to variable-length input sequences and requires an input selection algorithm to retain a consistent number of inputs with optimal coverage of the scene. As the pre-trained model for Long-LRM is not available, we directly use the numbers reported by the authors. 
We also compare to 3DGS~\cite{kerbl20233d}, which is an optimization-based approach that requires several minutes of optimization per scene. We generate a 3DGS representation using the same input images as our model, simply selecting every eighth image after removing all target images. 


As shown in Table~\ref{tab:results_main}, LVT$_{SH-rgba}$ outperforms both Long-LRM and 3DGS in PSNR, SSIM, and LPIPS on DL3DV. 
Our LVT$_{base}$ model, without spherical harmonics, offers a substantial boost in SSIM and LPIPS over Long-LRM, highlighting the effectiveness of variable-length input selection, as we do not require any heuristics to select a maximally informative fixed-length input set.

\begin{figure*}
  \includegraphics[width=0.9\linewidth]{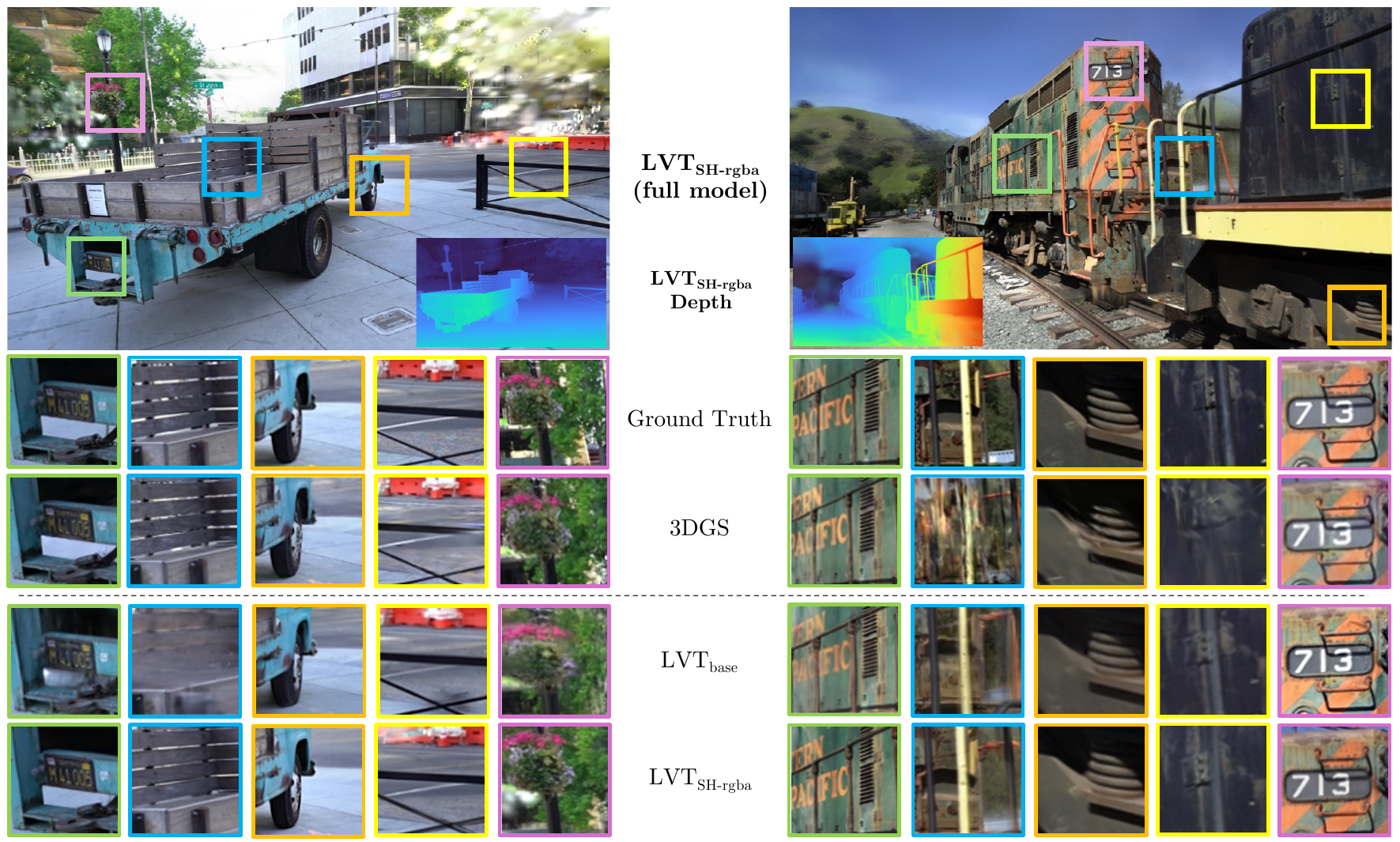}
 \vspace{-0.3cm}
  \caption{\textbf{Comparison of LVT models (zero-shot) on Tanks\&Temples dataset.} LVT trained on DL3DV generalizes well when tested zero-shot, with the full model achieving high-fidelity detail and structure reconstruction. Qualitatively, our LVT$_{SH-rgba}$ model preserves similar levels of detail as 3DGS, despite no per-scene optimization.}
  \label{fig:qual-tankstemples}
  \afterfigure
\end{figure*}

\begin{figure*}
  \includegraphics[width=0.9\linewidth]{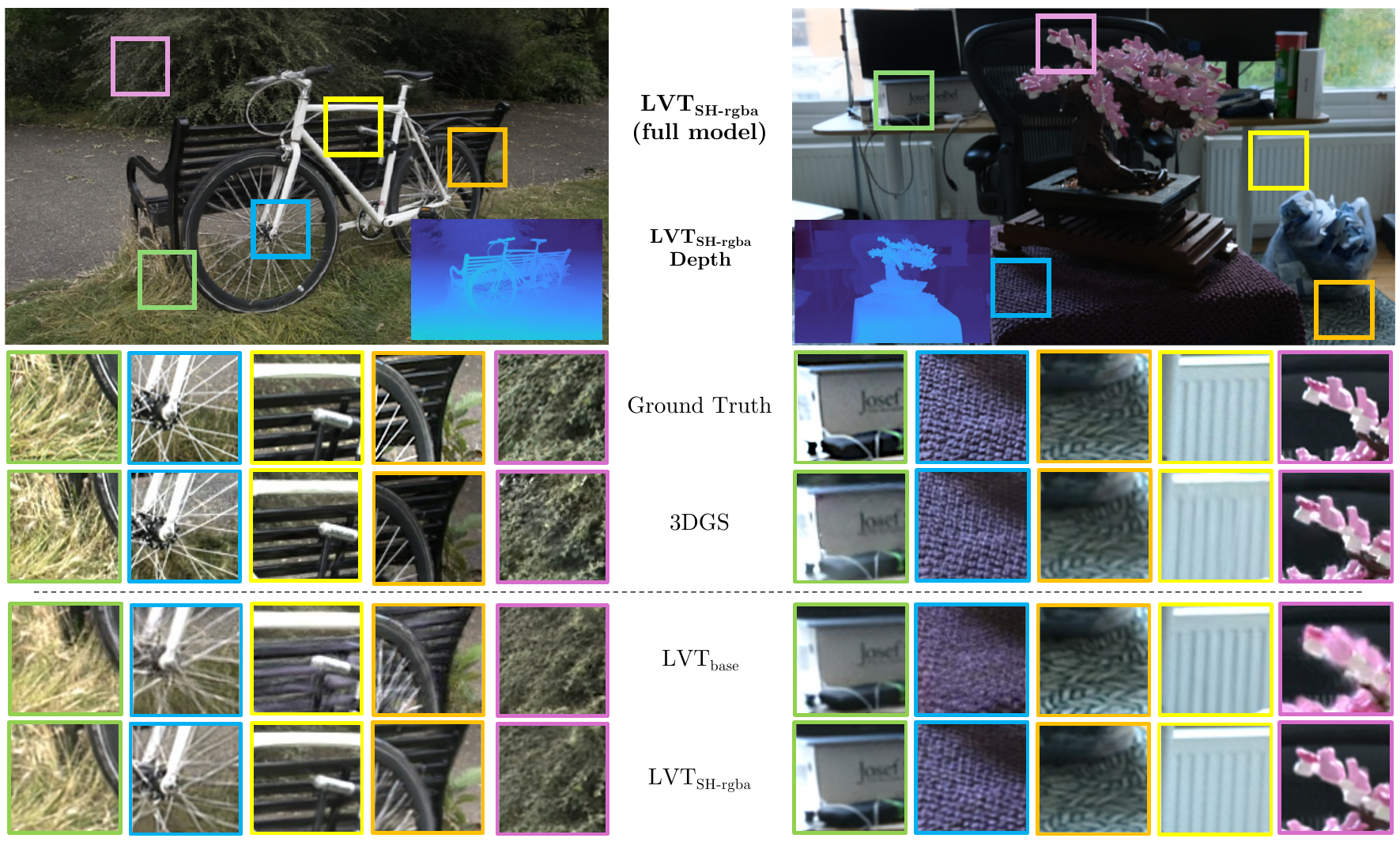}
  \vspace{-0.2cm}
  \caption{\textbf{Comparison of LVT (zero-shot) on Mip-NeRF360 dataset.} The camera trajectories in Mip-NeRF360 differ from the drone flight sequences used for training, yet our model is able to generalize to this dataset as it aggregates information in local neighborhoods. We note that 3DGS requires several minutes of optimization while our model performs an inference pass in seconds.}
  \label{fig:qual-mipnerf360}
  \afterfigure
\end{figure*}

We also evaluate our model trained on DL3DV on the \emph{train} and \emph{truck} examples from the Tanks\&Temples dataset, which we resize to $960\times540$ following Long-LRM (Figure \ref{fig:qual-tankstemples}). We run our model on 66 inputs for train and 55 for truck, longer than the sequence lengths that we use for training. Here, our LVT$_{base}$ model surpasses Long-LRM in all three metrics. We hypothesize that our local attention operation also helps with generalization given this different distribution of camera trajectories, as local segments of the Tanks\&Temples camera paths are more similar to those of DL3DV than the entire trajectory as a whole.
Our full model achieves better scores than our base model and, despite not being trained on this data, surpasses 3DGS in LPIPS, PSNR and SSIM.
We also evaluate our model on the MipNeRF360 dataset (Figure
\ref{fig:qual-mipnerf360}). Notably, Long-LRM reports inability to generalize to MipNeRF360. LVT outperforms 3DGS on this benchmark as well, and does so in a single inference pass. We benchmark the inference time of our model, which scales linearly with the number of inputs, in Supp.  Sec.~\ref{sec:supp-inference-time}.
\begin{table*}[t!]
    \centering
    \begin{tabular}{l c |c c c c |c c c c |c c c c } 
    \multicolumn{2}{c|}{} &
    \multicolumn{4}{c|}{\textit{PSNR} $\uparrow$} &
    \multicolumn{4}{c|}{\textit{SSIM} $\uparrow$} &
    \multicolumn{4}{c}{\textit{LPIPS} $\downarrow$} \\
    Number of Inputs & Complexity & 12 & 16 & 20 & 32 & 12 & 16 & 20 & 32 & 12 & 16 & 20 & 32 \\
    \midrule
    Global Attention & $\mathcal{O}(N^2)$ & 25.24 & 25.01 & 24.39 & 22.83 & 0.837 & 0.830 & 0.817 & 0.779 & 0.133 & 0.141 & 0.152 & 0.182 \\
    Local Attention & $\mathcal{O}(N)$ & {\textbf{25.32}} & {\textbf{25.12}} & {\textbf{24.69}} & {\textbf{23.31}} & {\textbf{0.846}} & {\textbf{0.842}} & {\textbf{0.834}} & {\textbf{0.807}} & {\textbf{0.130}} & {\textbf{0.135}} & {\textbf{0.141}} & {\textbf{0.167}} \\
    \bottomrule
    \end{tabular}
    \caption{\textbf{Comparison of local attention and global attention.} We train both models on 16 input views at $480\times270$ resolution. During inference, we change the number of input views and observe that LVT is more robust to differences in sequence length, while also exhibiting linear scaling behavior.}
    \label{tab:results_inputs}
    \aftertable
\end{table*}

\begin{table*}
    \centering
    \begin{tabular}{l|c c c c c c |c c c}
    \toprule
    Ablation & Mixed-Res. Train & Spatial Neighbor & Tokenize Cond. & Attention Cond. & SH & $\sigma$ Reg. & \textit{PSNR} $\uparrow$ & \textit{SSIM} $\uparrow$ & \textit{LPIPS} $\downarrow$ \\
    \midrule
    
    Initial Config. & \xmark & \xmark &  Intrinsics & Relative Pose & \xmark & \xmark & 22.02 & 0.774 & 0.196 \\
    Mixed-res. Train & \cmark & \xmark & Intrinsics & Relative Pose & \xmark & \xmark & 22.57 & 0.786 & 0.186 \\
    {Add Spatial} & {\xmark} & {\cmark} & {Intrinsics} & {Relative Pose} & {\xmark} & {\xmark} & {22.40} & {0.783} & {0.190} \\
    \midrule
    \multirow{3}{*}{\shortstack[l]{Conditioning\\Variants}}
    & \xmark & \xmark & Pl\"ucker & None & \xmark & \xmark & 20.59 & 0.696 & 0.260 \\
    & \xmark & \xmark & Intrinsics & World Pose & \xmark & \xmark & 21.44 & 0.745 & 0.220 \\
    & \xmark & \xmark & Intrinsics & Relative Pl\"ucker & \xmark & \xmark & 21.81 & 0.767 & 0.202 \\
    \midrule
    \multirow{2}{*}{\shortstack[l]{SH\\Variants}}
    & \xmark & \xmark & Intrinsics & Relative Pose & \cmark & \xmark  & 23.89 & 0.828 & 0.152 \\
    & \xmark & \xmark & Intrinsics & Relative Pose & \cmark & \cmark & \underline{24.69} & \underline{0.844} & \underline{0.136} \\
    \midrule
    Full Model & \cmark & {\cmark} & Intrinsics & Relative Pose & \cmark & \cmark & {\textbf{27.156}} & {\textbf{0.882}} & {\textbf{0.102}} \\

    \bottomrule
    \end{tabular}
    \caption{\textbf{Design ablations.} We train all configurations at $480\times270$ resolution. Our full model adds mixed-resolution training and regularized spherical harmonics to improve over our initial configuration, whereas we find that alternative conditioning variations lead to sub-optimal metrics.
    }
    \label{tab:results_ablation}
    \aftertable
\end{table*}

\subsection{Ablations}\label{sec:experiments-ablations}
We conduct ablations to validate the design choices and the effectiveness of our proposed Local View Transformer (LVT). We train all ablation models at $480\times270$ resolution with batch size 64.

\myparagraph{Generalization to varying sequence lengths.} We compare our LVT local attention to a model with full self-attention, following GS-LRM, but with otherwise identical training and model design to ours (Table \ref{tab:results_inputs}). We train both models on 16 inputs, and evaluate on 12, 16, 20, and 32 inputs to understand how sensitive the models are to varying numbers of inputs. Our model is both more robust to changes in the number of input views, while also being more scalable in the number of input views, with a linear rather than quadratic complexity attention module (see Sec.~\ref{sec:supp-inference-time}). {We hypothesize that this is because of the local, transformation-invariant design of LVT. By operating on \emph{local} neighborhoods using only \emph{relative} poses, any spatial reasoning learned is modular and independent of any global camera path. Relatedly, we hypothesize that the use of relative rather than absolute pose in our network simplifies the network's learning task. Using absolute poses requires learning over the combinatorial space of all key and query poses, whereas our approach constrains this problem by learning over the much smaller space of relative transformations between two nearby views.} However, we note that this version of our model is not fully invariant to the number of inputs - empirically, we find that while rendering to nearby cameras is accurate, incorrectly-placed distant Gaussians can end up interfering with other cameras in the sequence, so performance still degrades as sequence length increases.
This motivates our integration of multi-resolution and multi-length dataset mixing, which enables the accurate learning of 3D scene geometry from long sequences without requiring full-resolution training across all sequence lengths.

\myparagraph{Mixed-resolution training.} 
Our evaluations show that training with a mixed-resolution strategy leads to superior performance when evaluated across entire input sequences (whose lengths are scene-dependent), compared to models trained without this strategy.
While both models train on 16 inputs at $480\times270$ resolution the mixed-resolution model is also trained with longer sequences at lower $240\times135$ resolution. This enables it to develop a more comprehensive understanding of scene structure, that it then can transfer to higher resolutions. We report a qualitative comparison of the two configurations in the supplementary (Fig.~\ref{fig:mixres-ablation}).

\myparagraph{Spatial neighbor selection.} Since our training dataset consists of ordered image sequences, a simple way to construct the neighborhood graph is to take the images in sequential order, which we use for our initial configuration. However, we find that a spatial neighbor selection strategy, in which the neighboring images are selected based on camera center distance, outperforms the simple sequential image order strategy, and has the additional benefit of being agnostic to the order that input images are taken.

\myparagraph{Relative transformation conditioning.}\label{sec:rel-cond}
We conduct ablations on the method used to encode the relative transformations between the central view and its neighboring views within LVT. Table \ref{tab:results_ablation} compares different conditioning strategies. We find that using Pl\"ucker rays in the tokenization step, which encodes both camera intrinsics and extrinsics, but removing the conditioning to the attention module and thus removing information about relative transformations, degrades metrics. Encoding the local rays only during tokenization but directly using camera extrinsics relative to a world frame performs better than the previous variant but is still sub-optimal to our base model, which explicitly computes the relative transformations. Interestingly, we find that a straightforward, although more computationally expensive, encoding of the relative Pl\"ucker rays, e.g. applying $P_jP_{j'}^{-1}K_{j'}^{-1}$ to the pixel grid and adding to the attention conditioning, performs slightly worse than our base model, which directly encodes the relative quaternion and translation once per view and broadcasts that encoding across view tokens. {These experiments confirm the benefits of transformation-invariant relative embeddings, enabling generalization from shorter training sequences to longer inference sequences.}


\myparagraph{View-dependent splat parameters.}
As observed in Table~\ref{tab:results_main}, adding spherical harmonics on the color and opacity improves results over a model without view-dependent effects. We also ablate our opacity regularizer, which encourages the computed opacity from randomly sampled viewing directions to be zero. Adding this regularizer improves metrics, as the model without this regularization is prone to producing spurious opacity when seen from different viewing directions at inference time. Our best model combines these design choices, achieving the best metrics among the ablation variations.

\subsection{Limitations}

Our method achieves state-of-the-art performance on multiple benchmarks, and generalizes well to very long sequences with high fidelity,  with inference time scaling linearly with the input sequence length. However, it suffers from some limitations, which present an opportunity for future investigation. 

When input sequences contain many redundant, overlapping views, the model's inherently local attention mechanism may restrict communication between more distant, yet potentially relevant, views. Selecting a more spatially diverse subset of these input views, thereby increasing the typical separation between selected viewpoints, could potentially alleviate this issue. More complex neighbor selection strategies, that included both nearby and distant neighbors, could also allow for longer range communication. Alternatively, a hybrid approach could be explored, where LVT layers are interleaved with standard transformer layers.

Our model's per-pixel splat generation results in a dense, redundant representation when processing many high-resolution input views and we find that splat rasterization can become a significant bottleneck. Our use of view dependent opacity complicates the simple approach of culling almost-transparent splats, however, several previous works ~\cite{hanson_speedy,fast_lod_splats,niemeyer2025radsplat}, have introduced methods to efficiently render large 3DGS scene that could likely be applied to our model's outputs.

Finally, the use of view-dependent opacity can occasionally produce popping artifacts in areas with extreme textures (Supp. Fig~\ref{fig:failure-cases}). Increasing the regularization weight $\alpha$ can mitigate such artifacts, but this often comes at the cost of reduced rendering quality, indicating a trade-off.
\section{Conclusion}

We introduce the Local View Transformer (LVT), a new architecture for large-scale scene reconstruction and novel view synthesis. LVT's local attention mechanism allows scaling to many more input views than prior transformer-based methods.  Additionally, through the use of a relative, rather than global, camera pose conditioning mechanism, LVT improves generalization across long and varied camera trajectories.  
We further propose to decode to 3D Gaussian splats with both view-dependent color \emph{and} opacity, and demonstrate that this extension improves fidelity on thin structures and complex lighting effects.

Our experiments show that LVT achieves state-of-the-art results on a variety of benchmarks and demonstrates strong generalization capabilities. Furthermore, we perform comprehensive ablation studies confirming the efficacy of LVT's local attention, robustness to varying input sequence lengths, and benefits from mixed-resolution training, relative transformation conditioning, and view-dependent opacity with regularization. In summary, LVT offers an effective, scalable, and generalizable solution for high-fidelity, large-scale 3D scene reconstruction in a single forward pass.

\subsection*{Acknowledgments}

The authors would like to thank Stephen Lombardi, Clement Godard, Tiancheng Sun, and Yiming Wang for their invaluable feedback and discussions throughout the course of this research. We are also thankful to Peter Hedman, Daniel Duckworth, Ryan Overbeck, and Jason Lawrence for their valuable feedback and guidance for preparing this manuscript. Tooba Imtiaz was partly supported by NIH Graduate Research Fellowship under Grant No. 5U24CA264369-03.

\bibliographystyle{ACM-Reference-Format}
\bibliography{references}


\begin{thebibliography}{74}


\ifx \showCODEN    \undefined \def \showCODEN     #1{\unskip}     \fi
\ifx \showDOI      \undefined \def \showDOI       #1{#1}\fi
\ifx \showISBNx    \undefined \def \showISBNx     #1{\unskip}     \fi
\ifx \showISBNxiii \undefined \def \showISBNxiii  #1{\unskip}     \fi
\ifx \showISSN     \undefined \def \showISSN      #1{\unskip}     \fi
\ifx \showLCCN     \undefined \def \showLCCN      #1{\unskip}     \fi
\ifx \shownote     \undefined \def \shownote      #1{#1}          \fi
\ifx \showarticletitle \undefined \def \showarticletitle #1{#1}   \fi
\ifx \showURL      \undefined \def \showURL       {\relax}        \fi
\providecommand\bibfield[2]{#2}
\providecommand\bibinfo[2]{#2}
\providecommand\natexlab[1]{#1}
\providecommand\showeprint[2][]{arXiv:#2}

\bibitem[Agarwal et~al\mbox{.}(2009)]%
        {agarwal2009building}
\bibfield{author}{\bibinfo{person}{Sameer Agarwal}, \bibinfo{person}{Noah Snavely}, \bibinfo{person}{Ian Simon}, \bibinfo{person}{Steven~M. Seitz}, {and} \bibinfo{person}{Richard Szeliski}.} \bibinfo{year}{2009}\natexlab{}.
\newblock \showarticletitle{Building Rome in a Day}. In \bibinfo{booktitle}{\emph{Int. Conf. Comput. Vis.}} \bibinfo{pages}{72--79}.
\newblock


\bibitem[Aliev et~al\mbox{.}(2020)]%
        {aliev2020neural}
\bibfield{author}{\bibinfo{person}{Kara-Ali Aliev}, \bibinfo{person}{Artem Sevastopolsky}, \bibinfo{person}{Maria Kolos}, \bibinfo{person}{Dmitry Ulyanov}, {and} \bibinfo{person}{Victor Lempitsky}.} \bibinfo{year}{2020}\natexlab{}.
\newblock \showarticletitle{Neural point-based graphics}. In \bibinfo{booktitle}{\emph{Eur. Conf. Comput. Vis.}}
\newblock


\bibitem[Barron et~al\mbox{.}(2022)]%
        {barron2022mipnerf360}
\bibfield{author}{\bibinfo{person}{Jonathan~T. Barron}, \bibinfo{person}{Ben Mildenhall}, \bibinfo{person}{Dor Verbin}, \bibinfo{person}{Pratul~P. Srinivasan}, {and} \bibinfo{person}{Peter Hedman}.} \bibinfo{year}{2022}\natexlab{}.
\newblock \showarticletitle{Mip-NeRF 360: Unbounded Anti-Aliased Neural Radiance Fields}. In \bibinfo{booktitle}{\emph{IEEE Conf. Comput. Vis. Pattern Recog.}}
\newblock


\bibitem[Carion et~al\mbox{.}(2020)]%
        {carion2020end}
\bibfield{author}{\bibinfo{person}{Nicolas Carion}, \bibinfo{person}{Francisco Massa}, \bibinfo{person}{Gabriel Synnaeve}, \bibinfo{person}{Nicolas Usunier}, \bibinfo{person}{Alexander Kirillov}, {and} \bibinfo{person}{Sergey Zagoruyko}.} \bibinfo{year}{2020}\natexlab{}.
\newblock \showarticletitle{End-to-end object detection with transformers}. In \bibinfo{booktitle}{\emph{Eur. Conf. Comput. Vis.}} Springer, \bibinfo{pages}{213--229}.
\newblock


\bibitem[Caron et~al\mbox{.}(2021)]%
        {caron2021emerging}
\bibfield{author}{\bibinfo{person}{Mathilde Caron}, \bibinfo{person}{Hugo Touvron}, \bibinfo{person}{Ishan Misra}, \bibinfo{person}{Herv{\'e} J{\'e}gou}, \bibinfo{person}{Julien Mairal}, \bibinfo{person}{Piotr Bojanowski}, {and} \bibinfo{person}{Armand Joulin}.} \bibinfo{year}{2021}\natexlab{}.
\newblock \showarticletitle{Emerging properties in self-supervised vision transformers}. In \bibinfo{booktitle}{\emph{Int. Conf. Comput. Vis.}} \bibinfo{pages}{9650--9660}.
\newblock


\bibitem[Charatan et~al\mbox{.}(2024)]%
        {charatan2024pixelsplat}
\bibfield{author}{\bibinfo{person}{David Charatan}, \bibinfo{person}{Sizhe~Lester Li}, \bibinfo{person}{Andrea Tagliasacchi}, {and} \bibinfo{person}{Vincent Sitzmann}.} \bibinfo{year}{2024}\natexlab{}.
\newblock \showarticletitle{pixelsplat: 3d gaussian splats from image pairs for scalable generalizable 3d reconstruction}. In \bibinfo{booktitle}{\emph{IEEE Conf. Comput. Vis. Pattern Recog.}} \bibinfo{pages}{19457--19467}.
\newblock


\bibitem[Chaurasia et~al\mbox{.}(2013)]%
        {chaurasia2013depth}
\bibfield{author}{\bibinfo{person}{Gaurav Chaurasia}, \bibinfo{person}{Sylvain Duchene}, \bibinfo{person}{Olga Sorkine-Hornung}, {and} \bibinfo{person}{George Drettakis}.} \bibinfo{year}{2013}\natexlab{}.
\newblock \showarticletitle{Depth synthesis and local warps for plausible image-based navigation}.
\newblock \bibinfo{journal}{\emph{ACM Trans. Graph.}} \bibinfo{volume}{32}, \bibinfo{number}{3} (\bibinfo{year}{2013}), \bibinfo{pages}{1--12}.
\newblock


\bibitem[Chen et~al\mbox{.}(2021)]%
        {chen2021mvsnerf}
\bibfield{author}{\bibinfo{person}{Anpei Chen}, \bibinfo{person}{Zexiang Xu}, \bibinfo{person}{Fuqiang Zhao}, \bibinfo{person}{Xiaoshuai Zhang}, \bibinfo{person}{Fanbo Xiang}, \bibinfo{person}{Jingyi Yu}, {and} \bibinfo{person}{Hao Su}.} \bibinfo{year}{2021}\natexlab{}.
\newblock \showarticletitle{Mvsnerf: Fast generalizable radiance field reconstruction from multi-view stereo}. In \bibinfo{booktitle}{\emph{Int. Conf. Comput. Vis.}} \bibinfo{pages}{14124--14133}.
\newblock


\bibitem[Chen et~al\mbox{.}(2024)]%
        {chen2024mvsplat}
\bibfield{author}{\bibinfo{person}{Yuedong Chen}, \bibinfo{person}{Haofei Xu}, \bibinfo{person}{Chuanxia Zheng}, \bibinfo{person}{Bohan Zhuang}, \bibinfo{person}{Marc Pollefeys}, \bibinfo{person}{Andreas Geiger}, \bibinfo{person}{Tat-Jen Cham}, {and} \bibinfo{person}{Jianfei Cai}.} \bibinfo{year}{2024}\natexlab{}.
\newblock \showarticletitle{MVSplat: Efficient 3D Gaussian Splatting from Sparse Multi-View Images}. In \bibinfo{booktitle}{\emph{Eur. Conf. Comput. Vis.}}
\newblock


\bibitem[Dao et~al\mbox{.}(2022)]%
        {dao2022flashattention}
\bibfield{author}{\bibinfo{person}{Tri Dao}, \bibinfo{person}{Dan Fu}, \bibinfo{person}{Stefano Ermon}, \bibinfo{person}{Atri Rudra}, {and} \bibinfo{person}{Christopher R{\'e}}.} \bibinfo{year}{2022}\natexlab{}.
\newblock \showarticletitle{Flashattention: Fast and memory-efficient exact attention with io-awareness}. In \bibinfo{booktitle}{\emph{Adv. Neural Inform. Process. Syst.}}, Vol.~\bibinfo{volume}{35}. \bibinfo{pages}{16344--16359}.
\newblock


\bibitem[Dao and Gu(2024a)]%
        {mamba2}
\bibfield{author}{\bibinfo{person}{Tri Dao} {and} \bibinfo{person}{Albert Gu}.} \bibinfo{year}{2024}\natexlab{a}.
\newblock \showarticletitle{Transformers are {SSM}s: Generalized Models and Efficient Algorithms Through Structured State Space Duality}. In \bibinfo{booktitle}{\emph{Int. Conf. Machine Learning}}.
\newblock


\bibitem[Dao and Gu(2024b)]%
        {pmlr-v235-dao24a}
\bibfield{author}{\bibinfo{person}{Tri Dao} {and} \bibinfo{person}{Albert Gu}.} \bibinfo{year}{2024}\natexlab{b}.
\newblock \showarticletitle{Transformers are {SSM}s: Generalized Models and Efficient Algorithms Through Structured State Space Duality}. In \bibinfo{booktitle}{\emph{Proceedings of the 41st International Conference on Machine Learning}} \emph{(\bibinfo{series}{Proceedings of Machine Learning Research}, Vol.~\bibinfo{volume}{235})}. \bibinfo{publisher}{PMLR}, \bibinfo{pages}{10041--10071}.
\newblock
\urldef\tempurl%
\url{https://proceedings.mlr.press/v235/dao24a.html}
\showURL{%
\tempurl}


\bibitem[Devlin et~al\mbox{.}(2019)]%
        {devlin-etal-2019-bert}
\bibfield{author}{\bibinfo{person}{Jacob Devlin}, \bibinfo{person}{Ming-Wei Chang}, \bibinfo{person}{Kenton Lee}, {and} \bibinfo{person}{Kristina Toutanova}.} \bibinfo{year}{2019}\natexlab{}.
\newblock \showarticletitle{{BERT}: Pre-training of Deep Bidirectional Transformers for Language Understanding}. In \bibinfo{booktitle}{\emph{NAACL: Human Language Technologies, Volume 1 (Long and Short Papers)}}, \bibfield{editor}{\bibinfo{person}{Jill Burstein}, \bibinfo{person}{Christy Doran}, {and} \bibinfo{person}{Thamar Solorio}} (Eds.). \bibinfo{address}{Minneapolis, Minnesota}, \bibinfo{pages}{4171--4186}.
\newblock


\bibitem[Dosovitskiy et~al\mbox{.}(2021)]%
        {dosovitskiy2020vit}
\bibfield{author}{\bibinfo{person}{Alexey Dosovitskiy}, \bibinfo{person}{Lucas Beyer}, \bibinfo{person}{Alexander Kolesnikov}, \bibinfo{person}{Dirk Weissenborn}, \bibinfo{person}{Xiaohua Zhai}, \bibinfo{person}{Thomas Unterthiner}, \bibinfo{person}{Mostafa Dehghani}, \bibinfo{person}{Matthias Minderer}, \bibinfo{person}{Georg Heigold}, \bibinfo{person}{Sylvain Gelly}, \bibinfo{person}{Jakob Uszkoreit}, {and} \bibinfo{person}{Neil Houlsby}.} \bibinfo{year}{2021}\natexlab{}.
\newblock \showarticletitle{An Image is Worth 16x16 Words: Transformers for Image Recognition at Scale}. In \bibinfo{booktitle}{\emph{Int. Conf. Learn. Represent.}}
\newblock


\bibitem[Du et~al\mbox{.}(2023)]%
        {du2023learning}
\bibfield{author}{\bibinfo{person}{Yilun Du}, \bibinfo{person}{Cameron Smith}, \bibinfo{person}{Ayush Tewari}, {and} \bibinfo{person}{Vincent Sitzmann}.} \bibinfo{year}{2023}\natexlab{}.
\newblock \showarticletitle{Learning to render novel views from wide-baseline stereo pairs}. In \bibinfo{booktitle}{\emph{IEEE Conf. Comput. Vis. Pattern Recog.}} \bibinfo{pages}{4970--4980}.
\newblock


\bibitem[Flynn et~al\mbox{.}(2019)]%
        {flynn2019deepview}
\bibfield{author}{\bibinfo{person}{John Flynn}, \bibinfo{person}{Michael Broxton}, \bibinfo{person}{Paul Debevec}, \bibinfo{person}{Matthew DuVall}, \bibinfo{person}{Graham Fyffe}, \bibinfo{person}{Ryan Overbeck}, \bibinfo{person}{Noah Snavely}, {and} \bibinfo{person}{Richard Tucker}.} \bibinfo{year}{2019}\natexlab{}.
\newblock \showarticletitle{Deepview: View synthesis with learned gradient descent}. In \bibinfo{booktitle}{\emph{IEEE Conf. Comput. Vis. Pattern Recog.}} \bibinfo{pages}{2367--2376}.
\newblock


\bibitem[Flynn et~al\mbox{.}(2024)]%
        {flynn2024quark}
\bibfield{author}{\bibinfo{person}{John Flynn}, \bibinfo{person}{Michael Broxton}, \bibinfo{person}{Lukas Murmann}, \bibinfo{person}{Lucy Chai}, \bibinfo{person}{Matthew DuVall}, \bibinfo{person}{Cl{\'e}ment Godard}, \bibinfo{person}{Kathryn Heal}, \bibinfo{person}{Srinivas Kaza}, \bibinfo{person}{Stephen Lombardi}, \bibinfo{person}{Xuan Luo}, {et~al\mbox{.}}} \bibinfo{year}{2024}\natexlab{}.
\newblock \showarticletitle{Quark: Real-time, High-resolution, and General Neural View Synthesis}.
\newblock \bibinfo{journal}{\emph{ACM Trans. Graph.}} \bibinfo{volume}{43}, \bibinfo{number}{6} (\bibinfo{year}{2024}), \bibinfo{pages}{1--20}.
\newblock


\bibitem[Flynn et~al\mbox{.}(2016)]%
        {flynn2016deepstereo}
\bibfield{author}{\bibinfo{person}{John Flynn}, \bibinfo{person}{Ivan Neulander}, \bibinfo{person}{James Philbin}, {and} \bibinfo{person}{Noah Snavely}.} \bibinfo{year}{2016}\natexlab{}.
\newblock \showarticletitle{Deepstereo: Learning to predict new views from the world's imagery}. In \bibinfo{booktitle}{\emph{IEEE Conf. Comput. Vis. Pattern Recog.}} \bibinfo{pages}{5515--5524}.
\newblock


\bibitem[Fridovich-Keil et~al\mbox{.}(2022)]%
        {fridovich2022plenoxels}
\bibfield{author}{\bibinfo{person}{Sara Fridovich-Keil}, \bibinfo{person}{Alex Yu}, \bibinfo{person}{Matthew Tancik}, \bibinfo{person}{Qinhong Chen}, \bibinfo{person}{Benjamin Recht}, {and} \bibinfo{person}{Angjoo Kanazawa}.} \bibinfo{year}{2022}\natexlab{}.
\newblock \showarticletitle{Plenoxels: Radiance fields without neural networks}. In \bibinfo{booktitle}{\emph{IEEE Conf. Comput. Vis. Pattern Recog.}} \bibinfo{pages}{5501--5510}.
\newblock


\bibitem[Hanson et~al\mbox{.}(2025)]%
        {hanson_speedy}
\bibfield{author}{\bibinfo{person}{Alex Hanson}, \bibinfo{person}{Allen Tu}, \bibinfo{person}{Geng Lin}, \bibinfo{person}{Vasu Singla}, \bibinfo{person}{Matthias Zwicker}, {and} \bibinfo{person}{Tom Goldstein}.} \bibinfo{year}{2025}\natexlab{}.
\newblock \showarticletitle{Speedy-Splat: Fast 3D Gaussian Splatting with Sparse Pixels and Sparse Primitives}. In \bibinfo{booktitle}{\emph{IEEE Conf. Comput. Vis. Pattern Recog.}}
\newblock


\bibitem[Heitz et~al\mbox{.}(2015)]%
        {heitz2015sggx}
\bibfield{author}{\bibinfo{person}{Eric Heitz}, \bibinfo{person}{Jonathan Dupuy}, \bibinfo{person}{Cyril Crassin}, {and} \bibinfo{person}{Carsten Dachsbacher}.} \bibinfo{year}{2015}\natexlab{}.
\newblock \showarticletitle{The SGGX microflake distribution}.
\newblock \bibinfo{journal}{\emph{ACM Trans. Graph.}} \bibinfo{volume}{34}, \bibinfo{number}{4} (\bibinfo{year}{2015}), \bibinfo{pages}{1--11}.
\newblock


\bibitem[Heo et~al\mbox{.}(2024)]%
        {heo2024rotary}
\bibfield{author}{\bibinfo{person}{Byeongho Heo}, \bibinfo{person}{Song Park}, \bibinfo{person}{Dongyoon Han}, {and} \bibinfo{person}{Sangdoo Yun}.} \bibinfo{year}{2024}\natexlab{}.
\newblock \showarticletitle{Rotary position embedding for vision transformer}. In \bibinfo{booktitle}{\emph{Eur. Conf. Comput. Vis.}} Springer, \bibinfo{pages}{289--305}.
\newblock


\bibitem[Hong et~al\mbox{.}(2024)]%
        {hong2023lrm}
\bibfield{author}{\bibinfo{person}{Yicong Hong}, \bibinfo{person}{Kai Zhang}, \bibinfo{person}{Jiuxiang Gu}, \bibinfo{person}{Sai Bi}, \bibinfo{person}{Yang Zhou}, \bibinfo{person}{Difan Liu}, \bibinfo{person}{Feng Liu}, \bibinfo{person}{Kalyan Sunkavalli}, \bibinfo{person}{Trung Bui}, {and} \bibinfo{person}{Hao Tan}.} \bibinfo{year}{2024}\natexlab{}.
\newblock \showarticletitle{Lrm: Large reconstruction model for single image to 3d}.
\newblock \bibinfo{journal}{\emph{Int. Conf. Learn. Represent.}} (\bibinfo{year}{2024}).
\newblock


\bibitem[Jin et~al\mbox{.}(2025)]%
        {jin2025lvsm}
\bibfield{author}{\bibinfo{person}{Haian Jin}, \bibinfo{person}{Hanwen Jiang}, \bibinfo{person}{Hao Tan}, \bibinfo{person}{Kai Zhang}, \bibinfo{person}{Sai Bi}, \bibinfo{person}{Tianyuan Zhang}, \bibinfo{person}{Fujun Luan}, \bibinfo{person}{Noah Snavely}, {and} \bibinfo{person}{Zexiang Xu}.} \bibinfo{year}{2025}\natexlab{}.
\newblock \showarticletitle{LVSM: A Large View Synthesis Model with Minimal 3D Inductive Bias}. In \bibinfo{booktitle}{\emph{Int. Conf. Learn. Represent.}}
\newblock


\bibitem[Johari et~al\mbox{.}(2022)]%
        {johari2022geonerf}
\bibfield{author}{\bibinfo{person}{Mohammad~Mahdi Johari}, \bibinfo{person}{Yann Lepoittevin}, {and} \bibinfo{person}{Fran{\c{c}}ois Fleuret}.} \bibinfo{year}{2022}\natexlab{}.
\newblock \showarticletitle{Geonerf: Generalizing nerf with geometry priors}. In \bibinfo{booktitle}{\emph{IEEE Conf. Comput. Vis. Pattern Recog.}} \bibinfo{pages}{18365--18375}.
\newblock


\bibitem[Kerbl et~al\mbox{.}(2023)]%
        {kerbl20233d}
\bibfield{author}{\bibinfo{person}{Bernhard Kerbl}, \bibinfo{person}{Georgios Kopanas}, \bibinfo{person}{Thomas Leimk{\"u}hler}, {and} \bibinfo{person}{George Drettakis}.} \bibinfo{year}{2023}\natexlab{}.
\newblock \showarticletitle{3d gaussian splatting for real-time radiance field rendering.}
\newblock \bibinfo{journal}{\emph{ACM Trans. Graph.}} \bibinfo{volume}{42}, \bibinfo{number}{4} (\bibinfo{year}{2023}), \bibinfo{pages}{139--1}.
\newblock


\bibitem[Knapitsch et~al\mbox{.}(2017)]%
        {knapitsch2017tanks}
\bibfield{author}{\bibinfo{person}{Arno Knapitsch}, \bibinfo{person}{Jaesik Park}, \bibinfo{person}{Qian-Yi Zhou}, {and} \bibinfo{person}{Vladlen Koltun}.} \bibinfo{year}{2017}\natexlab{}.
\newblock \showarticletitle{Tanks and temples: Benchmarking large-scale scene reconstruction}.
\newblock \bibinfo{journal}{\emph{ACM Trans. Graph.}} \bibinfo{volume}{36}, \bibinfo{number}{4} (\bibinfo{year}{2017}), \bibinfo{pages}{1--13}.
\newblock


\bibitem[Kong et~al\mbox{.}(2024)]%
        {kong2024eschernet}
\bibfield{author}{\bibinfo{person}{Xin Kong}, \bibinfo{person}{Shikun Liu}, \bibinfo{person}{Xiaoyang Lyu}, \bibinfo{person}{Marwan Taher}, \bibinfo{person}{Xiaojuan Qi}, {and} \bibinfo{person}{Andrew~J Davison}.} \bibinfo{year}{2024}\natexlab{}.
\newblock \showarticletitle{Eschernet: A generative model for scalable view synthesis}. In \bibinfo{booktitle}{\emph{IEEE Conf. Comput. Vis. Pattern Recog.}} \bibinfo{pages}{9503--9513}.
\newblock


\bibitem[Kutulakos and Seitz(2000)]%
        {kutulakos2000theory}
\bibfield{author}{\bibinfo{person}{Kiriakos~N. Kutulakos} {and} \bibinfo{person}{Steven~M. Seitz}.} \bibinfo{year}{2000}\natexlab{}.
\newblock \showarticletitle{A Theory of Shape by Space Carving}.
\newblock \bibinfo{journal}{\emph{Int. J. Comput. Vis.}} \bibinfo{volume}{38}, \bibinfo{number}{3} (\bibinfo{year}{2000}), \bibinfo{pages}{199--218}.
\newblock
\urldef\tempurl%
\url{https://doi.org/10.1023/A:1008191222954}
\showDOI{\tempurl}


\bibitem[LeCun et~al\mbox{.}(2015)]%
        {lecun2015deep}
\bibfield{author}{\bibinfo{person}{Yann LeCun}, \bibinfo{person}{Yoshua Bengio}, {and} \bibinfo{person}{Geoffrey Hinton}.} \bibinfo{year}{2015}\natexlab{}.
\newblock \showarticletitle{Deep learning}.
\newblock \bibinfo{journal}{\emph{nature}} \bibinfo{volume}{521}, \bibinfo{number}{7553} (\bibinfo{year}{2015}), \bibinfo{pages}{436--444}.
\newblock


\bibitem[Li et~al\mbox{.}(2023)]%
        {li2023instant3d}
\bibfield{author}{\bibinfo{person}{Jiahao Li}, \bibinfo{person}{Hao Tan}, \bibinfo{person}{Kai Zhang}, \bibinfo{person}{Zexiang Xu}, \bibinfo{person}{Fujun Luan}, \bibinfo{person}{Yinghao Xu}, \bibinfo{person}{Yicong Hong}, \bibinfo{person}{Kalyan Sunkavalli}, \bibinfo{person}{Greg Shakhnarovich}, {and} \bibinfo{person}{Sai Bi}.} \bibinfo{year}{2023}\natexlab{}.
\newblock \showarticletitle{Instant3d: Fast text-to-3d with sparse-view generation and large reconstruction model}.
\newblock \bibinfo{journal}{\emph{arXiv preprint arXiv:2311.06214}} (\bibinfo{year}{2023}).
\newblock


\bibitem[Li et~al\mbox{.}(2025)]%
        {li2025cameras}
\bibfield{author}{\bibinfo{person}{Ruilong Li}, \bibinfo{person}{Brent Yi}, \bibinfo{person}{Junchen Liu}, \bibinfo{person}{Hang Gao}, \bibinfo{person}{Yi Ma}, {and} \bibinfo{person}{Angjoo Kanazawa}.} \bibinfo{year}{2025}\natexlab{}.
\newblock \showarticletitle{Cameras as Relative Positional Encoding}.
\newblock \bibinfo{journal}{\emph{arXiv preprint arXiv:2507.10496}} (\bibinfo{year}{2025}).
\newblock


\bibitem[Lin et~al\mbox{.}(2020)]%
        {lin2020deep}
\bibfield{author}{\bibinfo{person}{Kai-En Lin}, \bibinfo{person}{Zexiang Xu}, \bibinfo{person}{Ben Mildenhall}, \bibinfo{person}{Pratul~P Srinivasan}, \bibinfo{person}{Yannick Hold-Geoffroy}, \bibinfo{person}{Stephen DiVerdi}, \bibinfo{person}{Qi Sun}, \bibinfo{person}{Kalyan Sunkavalli}, {and} \bibinfo{person}{Ravi Ramamoorthi}.} \bibinfo{year}{2020}\natexlab{}.
\newblock \showarticletitle{Deep multi depth panoramas for view synthesis}. In \bibinfo{booktitle}{\emph{Eur. Conf. Comput. Vis.}} Springer, \bibinfo{pages}{328--344}.
\newblock


\bibitem[Ling et~al\mbox{.}(2024)]%
        {ling2024dl3dv}
\bibfield{author}{\bibinfo{person}{Lu Ling}, \bibinfo{person}{Yichen Sheng}, \bibinfo{person}{Zhi Tu}, \bibinfo{person}{Wentian Zhao}, \bibinfo{person}{Cheng Xin}, \bibinfo{person}{Kun Wan}, \bibinfo{person}{Lantao Yu}, \bibinfo{person}{Qianyu Guo}, \bibinfo{person}{Zixun Yu}, \bibinfo{person}{Yawen Lu}, {et~al\mbox{.}}} \bibinfo{year}{2024}\natexlab{}.
\newblock \showarticletitle{Dl3dv-10k: A large-scale scene dataset for deep learning-based 3d vision}. In \bibinfo{booktitle}{\emph{IEEE Conf. Comput. Vis. Pattern Recog.}} \bibinfo{pages}{22160--22169}.
\newblock


\bibitem[Liu et~al\mbox{.}(2020)]%
        {liu2020neural}
\bibfield{author}{\bibinfo{person}{Lingjie Liu}, \bibinfo{person}{Jiatao Gu}, \bibinfo{person}{Kyaw Zaw~Lin}, \bibinfo{person}{Tat-Seng Chua}, {and} \bibinfo{person}{Christian Theobalt}.} \bibinfo{year}{2020}\natexlab{}.
\newblock \showarticletitle{Neural sparse voxel fields}. In \bibinfo{booktitle}{\emph{Adv. Neural Inform. Process. Syst.}}, Vol.~\bibinfo{volume}{33}. \bibinfo{pages}{15651--15663}.
\newblock


\bibitem[Mildenhall et~al\mbox{.}(2019)]%
        {mildenhall2019llff}
\bibfield{author}{\bibinfo{person}{Ben Mildenhall}, \bibinfo{person}{Pratul~P. Srinivasan}, \bibinfo{person}{Rodrigo Ortiz-Cayon}, \bibinfo{person}{Nima~Khademi Kalantari}, \bibinfo{person}{Ravi Ramamoorthi}, \bibinfo{person}{Ren Ng}, {and} \bibinfo{person}{Abhishek Kar}.} \bibinfo{year}{2019}\natexlab{}.
\newblock \showarticletitle{Local Light Field Fusion: Practical View Synthesis with Prescriptive Sampling Guidelines}.
\newblock \bibinfo{journal}{\emph{ACM Trans. Graph.}} (\bibinfo{year}{2019}).
\newblock


\bibitem[Mildenhall et~al\mbox{.}(2020)]%
        {mildenhall2021nerf}
\bibfield{author}{\bibinfo{person}{Ben Mildenhall}, \bibinfo{person}{Pratul~P Srinivasan}, \bibinfo{person}{Matthew Tancik}, \bibinfo{person}{Jonathan~T Barron}, \bibinfo{person}{Ravi Ramamoorthi}, {and} \bibinfo{person}{Ren Ng}.} \bibinfo{year}{2020}\natexlab{}.
\newblock \showarticletitle{Nerf: Representing scenes as neural radiance fields for view synthesis}. In \bibinfo{booktitle}{\emph{Eur. Conf. Comput. Vis.}}
\newblock


\bibitem[Milef et~al\mbox{.}(2025)]%
        {fast_lod_splats}
\bibfield{author}{\bibinfo{person}{N. Milef}, \bibinfo{person}{D. Seyb}, \bibinfo{person}{T. Keeler}, \bibinfo{person}{T. Nguyen‐Phuoc}, \bibinfo{person}{A. Božič}, \bibinfo{person}{S. Kondguli}, {and} \bibinfo{person}{C. Marshall}.} \bibinfo{year}{2025}\natexlab{}.
\newblock \showarticletitle{Learning Fast 3D Gaussian Splatting Rendering using Continuous Level of Detail}.
\newblock \bibinfo{journal}{\emph{Computer Graphics Forum}} (\bibinfo{date}{04} \bibinfo{year}{2025}).
\newblock
\urldef\tempurl%
\url{https://doi.org/10.1111/cgf.70069}
\showDOI{\tempurl}


\bibitem[Miyato et~al\mbox{.}(2024)]%
        {miyato2023gta}
\bibfield{author}{\bibinfo{person}{Takeru Miyato}, \bibinfo{person}{Bernhard Jaeger}, \bibinfo{person}{Max Welling}, {and} \bibinfo{person}{Andreas Geiger}.} \bibinfo{year}{2024}\natexlab{}.
\newblock \showarticletitle{Gta: A geometry-aware attention mechanism for multi-view transformers}. In \bibinfo{booktitle}{\emph{Int. Conf. Learn. Represent.}}
\newblock


\bibitem[M{\"u}ller et~al\mbox{.}(2022)]%
        {muller2022instant}
\bibfield{author}{\bibinfo{person}{Thomas M{\"u}ller}, \bibinfo{person}{Alex Evans}, \bibinfo{person}{Christoph Schied}, {and} \bibinfo{person}{Alexander Keller}.} \bibinfo{year}{2022}\natexlab{}.
\newblock \showarticletitle{Instant neural graphics primitives with a multiresolution hash encoding}.
\newblock \bibinfo{journal}{\emph{ACM Trans. Graph.}} \bibinfo{volume}{41}, \bibinfo{number}{4} (\bibinfo{year}{2022}), \bibinfo{pages}{1--15}.
\newblock


\bibitem[Niemeyer et~al\mbox{.}(2025)]%
        {niemeyer2025radsplat}
\bibfield{author}{\bibinfo{person}{Michael Niemeyer}, \bibinfo{person}{Fabian Manhardt}, \bibinfo{person}{Marie-Julie Rakotosaona}, \bibinfo{person}{Michael Oechsle}, \bibinfo{person}{Daniel Duckworth}, \bibinfo{person}{Rama Gosula}, \bibinfo{person}{Keisuke Tateno}, \bibinfo{person}{John Bates}, \bibinfo{person}{Dominik Kaeser}, {and} \bibinfo{person}{Federico Tombari}.} \bibinfo{year}{2025}\natexlab{}.
\newblock \showarticletitle{RadSplat: Radiance Field-Informed Gaussian Splatting for Robust Real-Time Rendering with 900+ FPS}. In \bibinfo{booktitle}{\emph{Int. Conf. on 3D Vision}}.
\newblock


\bibitem[Nowak et~al\mbox{.}(2025)]%
        {nowak2025vod}
\bibfield{author}{\bibinfo{person}{Mateusz Nowak}, \bibinfo{person}{Wojciech Jarosz}, {and} \bibinfo{person}{Peter Chin}.} \bibinfo{year}{2025}\natexlab{}.
\newblock \showarticletitle{VoD-3DGS: View-opacity-Dependent 3D Gaussian Splatting}.
\newblock \bibinfo{journal}{\emph{arXiv preprint arXiv:2501.17978}} (\bibinfo{year}{2025}).
\newblock


\bibitem[Penner and Zhang(2017)]%
        {penner2017soft}
\bibfield{author}{\bibinfo{person}{Eric Penner} {and} \bibinfo{person}{Li Zhang}.} \bibinfo{year}{2017}\natexlab{}.
\newblock \showarticletitle{Soft 3d reconstruction for view synthesis}.
\newblock \bibinfo{journal}{\emph{ACM Trans. Graph.}} \bibinfo{volume}{36}, \bibinfo{number}{6} (\bibinfo{year}{2017}), \bibinfo{pages}{1--11}.
\newblock


\bibitem[Pl\"ucker(1865)]%
        {plucker1865xvii}
\bibfield{author}{\bibinfo{person}{Julius Pl\"ucker}.} \bibinfo{year}{1865}\natexlab{}.
\newblock \showarticletitle{Xvii. on a new geometry of space}.
\newblock \bibinfo{journal}{\emph{Philosophical Transactions of the Royal Society of London}} \bibinfo{number}{155} (\bibinfo{year}{1865}), \bibinfo{pages}{725--791}.
\newblock


\bibitem[Radford et~al\mbox{.}(2019)]%
        {radford2019language}
\bibfield{author}{\bibinfo{person}{Alec Radford}, \bibinfo{person}{Jeff Wu}, \bibinfo{person}{Rewon Child}, \bibinfo{person}{David Luan}, \bibinfo{person}{Dario Amodei}, {and} \bibinfo{person}{Ilya Sutskever}.} \bibinfo{year}{2019}\natexlab{}.
\newblock \showarticletitle{Language Models are Unsupervised Multitask Learners}.
\newblock  (\bibinfo{year}{2019}).
\newblock


\bibitem[Ramamoorthi and Hanrahan(2001)]%
        {ramamoorthi_sh}
\bibfield{author}{\bibinfo{person}{Ravi Ramamoorthi} {and} \bibinfo{person}{Pat Hanrahan}.} \bibinfo{year}{2001}\natexlab{}.
\newblock \showarticletitle{An efficient representation for irradiance environment maps}. In \bibinfo{booktitle}{\emph{Computer Graphics and Interactive Techniques}} \emph{(\bibinfo{series}{SIGGRAPH '01})}. \bibinfo{publisher}{Association for Computing Machinery}, \bibinfo{address}{New York, NY, USA}, \bibinfo{pages}{497–500}.
\newblock
\showISBNx{158113374X}
\urldef\tempurl%
\url{https://doi.org/10.1145/383259.383317}
\showDOI{\tempurl}


\bibitem[Ranftl et~al\mbox{.}(2021)]%
        {ranftl2021vision}
\bibfield{author}{\bibinfo{person}{Ren{\'e} Ranftl}, \bibinfo{person}{Alexey Bochkovskiy}, {and} \bibinfo{person}{Vladlen Koltun}.} \bibinfo{year}{2021}\natexlab{}.
\newblock \showarticletitle{Vision transformers for dense prediction}. In \bibinfo{booktitle}{\emph{Int. Conf. Comput. Vis.}} \bibinfo{pages}{12179--12188}.
\newblock


\bibitem[Sch{\"o}nberger et~al\mbox{.}(2016)]%
        {schonberger2016pixelwise}
\bibfield{author}{\bibinfo{person}{Johannes~L Sch{\"o}nberger}, \bibinfo{person}{Enliang Zheng}, \bibinfo{person}{Jan-Michael Frahm}, {and} \bibinfo{person}{Marc Pollefeys}.} \bibinfo{year}{2016}\natexlab{}.
\newblock \showarticletitle{Pixelwise view selection for unstructured multi-view stereo}. In \bibinfo{booktitle}{\emph{Eur. Conf. Comput. Vis.}} Springer, \bibinfo{pages}{501--518}.
\newblock


\bibitem[Seitz and Dyer(1999)]%
        {seitz1999photoconsistency}
\bibfield{author}{\bibinfo{person}{Steven~M. Seitz} {and} \bibinfo{person}{Charles~R. Dyer}.} \bibinfo{year}{1999}\natexlab{}.
\newblock \showarticletitle{Photorealistic Scene Reconstruction by Voxel Coloring}.
\newblock \bibinfo{journal}{\emph{Int. J. Comput. Vis.}} \bibinfo{volume}{35}, \bibinfo{number}{2} (\bibinfo{year}{1999}), \bibinfo{pages}{151--173}.
\newblock
\urldef\tempurl%
\url{https://doi.org/10.1023/A:1008176507526}
\showDOI{\tempurl}


\bibitem[Simonyan and Zisserman(2014)]%
        {simonyan2014very}
\bibfield{author}{\bibinfo{person}{Karen Simonyan} {and} \bibinfo{person}{Andrew Zisserman}.} \bibinfo{year}{2014}\natexlab{}.
\newblock \showarticletitle{Very deep convolutional networks for large-scale image recognition}.
\newblock \bibinfo{journal}{\emph{arXiv preprint arXiv:1409.1556}} (\bibinfo{year}{2014}).
\newblock


\bibitem[Sloan et~al\mbox{.}(2023)]%
        {sloan2023precomputed}
\bibfield{author}{\bibinfo{person}{Peter-Pike Sloan}, \bibinfo{person}{Jan Kautz}, {and} \bibinfo{person}{John Snyder}.} \bibinfo{year}{2023}\natexlab{}.
\newblock \showarticletitle{Precomputed radiance transfer for real-time rendering in dynamic, low-frequency lighting environments}.
\newblock In \bibinfo{booktitle}{\emph{Seminal Graphics Papers: Pushing the Boundaries, Volume 2}}. \bibinfo{pages}{339--348}.
\newblock


\bibitem[Su et~al\mbox{.}(2024)]%
        {su2024roformer}
\bibfield{author}{\bibinfo{person}{Jianlin Su}, \bibinfo{person}{Murtadha Ahmed}, \bibinfo{person}{Yu Lu}, \bibinfo{person}{Shengfeng Pan}, \bibinfo{person}{Wen Bo}, {and} \bibinfo{person}{Yunfeng Liu}.} \bibinfo{year}{2024}\natexlab{}.
\newblock \showarticletitle{Roformer: Enhanced transformer with rotary position embedding}.
\newblock \bibinfo{journal}{\emph{Neurocomputing}}  \bibinfo{volume}{568} (\bibinfo{year}{2024}), \bibinfo{pages}{127063}.
\newblock


\bibitem[Suhail et~al\mbox{.}(2022)]%
        {suhail2022generalizable}
\bibfield{author}{\bibinfo{person}{Mohammed Suhail}, \bibinfo{person}{Carlos Esteves}, \bibinfo{person}{Leonid Sigal}, {and} \bibinfo{person}{Ameesh Makadia}.} \bibinfo{year}{2022}\natexlab{}.
\newblock \showarticletitle{Generalizable patch-based neural rendering}. In \bibinfo{booktitle}{\emph{Eur. Conf. Comput. Vis.}} Springer, \bibinfo{pages}{156--174}.
\newblock


\bibitem[Tang et~al\mbox{.}(2024)]%
        {tang2024lgm}
\bibfield{author}{\bibinfo{person}{Jiaxiang Tang}, \bibinfo{person}{Zhaoxi Chen}, \bibinfo{person}{Xiaokang Chen}, \bibinfo{person}{Tengfei Wang}, \bibinfo{person}{Gang Zeng}, {and} \bibinfo{person}{Ziwei Liu}.} \bibinfo{year}{2024}\natexlab{}.
\newblock \showarticletitle{Lgm: Large multi-view gaussian model for high-resolution 3d content creation}. In \bibinfo{booktitle}{\emph{Eur. Conf. Comput. Vis.}} Springer, \bibinfo{pages}{1--18}.
\newblock


\bibitem[Vaswani et~al\mbox{.}(2017)]%
        {vaswani2017attention}
\bibfield{author}{\bibinfo{person}{Ashish Vaswani}, \bibinfo{person}{Noam Shazeer}, \bibinfo{person}{Niki Parmar}, \bibinfo{person}{Jakob Uszkoreit}, \bibinfo{person}{Llion Jones}, \bibinfo{person}{Aidan~N Gomez}, \bibinfo{person}{{\L}ukasz Kaiser}, {and} \bibinfo{person}{Illia Polosukhin}.} \bibinfo{year}{2017}\natexlab{}.
\newblock \showarticletitle{Attention is all you need}.
\newblock \bibinfo{journal}{\emph{Adv. Neural Inform. Process. Syst.}}  \bibinfo{volume}{30} (\bibinfo{year}{2017}).
\newblock


\bibitem[Verbin et~al\mbox{.}(2022)]%
        {verbin2022ref}
\bibfield{author}{\bibinfo{person}{Dor Verbin}, \bibinfo{person}{Peter Hedman}, \bibinfo{person}{Ben Mildenhall}, \bibinfo{person}{Todd Zickler}, \bibinfo{person}{Jonathan~T Barron}, {and} \bibinfo{person}{Pratul~P Srinivasan}.} \bibinfo{year}{2022}\natexlab{}.
\newblock \showarticletitle{Ref-nerf: Structured view-dependent appearance for neural radiance fields}. In \bibinfo{booktitle}{\emph{IEEE Conf. Comput. Vis. Pattern Recog.}} IEEE, \bibinfo{pages}{5481--5490}.
\newblock


\bibitem[Wang and Agapito(2025)]%
        {wang20243dreconstructionspatialmemory}
\bibfield{author}{\bibinfo{person}{Hengyi Wang} {and} \bibinfo{person}{Lourdes Agapito}.} \bibinfo{year}{2025}\natexlab{}.
\newblock \showarticletitle{3D Reconstruction with Spatial Memory}. In \bibinfo{booktitle}{\emph{Int. Conf. on 3D Vision}}.
\newblock


\bibitem[Wang et~al\mbox{.}(2024b)]%
        {wang2023pf}
\bibfield{author}{\bibinfo{person}{Peng Wang}, \bibinfo{person}{Hao Tan}, \bibinfo{person}{Sai Bi}, \bibinfo{person}{Yinghao Xu}, \bibinfo{person}{Fujun Luan}, \bibinfo{person}{Kalyan Sunkavalli}, \bibinfo{person}{Wenping Wang}, \bibinfo{person}{Zexiang Xu}, {and} \bibinfo{person}{Kai Zhang}.} \bibinfo{year}{2024}\natexlab{b}.
\newblock \showarticletitle{Pf-lrm: Pose-free large reconstruction model for joint pose and shape prediction}. In \bibinfo{booktitle}{\emph{Int. Conf. Learn. Represent.}}
\newblock


\bibitem[Wang et~al\mbox{.}(2021)]%
        {wang2021ibrnet}
\bibfield{author}{\bibinfo{person}{Qianqian Wang}, \bibinfo{person}{Zhicheng Wang}, \bibinfo{person}{Kyle Genova}, \bibinfo{person}{Pratul~P Srinivasan}, \bibinfo{person}{Howard Zhou}, \bibinfo{person}{Jonathan~T Barron}, \bibinfo{person}{Ricardo Martin-Brualla}, \bibinfo{person}{Noah Snavely}, {and} \bibinfo{person}{Thomas Funkhouser}.} \bibinfo{year}{2021}\natexlab{}.
\newblock \showarticletitle{Ibrnet: Learning multi-view image-based rendering}. In \bibinfo{booktitle}{\emph{IEEE Conf. Comput. Vis. Pattern Recog.}} \bibinfo{pages}{4690--4699}.
\newblock


\bibitem[Wang* et~al\mbox{.}(2025)]%
        {wang2025continuous3dperceptionmodel}
\bibfield{author}{\bibinfo{person}{Qianqian Wang*}, \bibinfo{person}{Yifei Zhang*}, \bibinfo{person}{Aleksander Holynski}, \bibinfo{person}{Alexei~A. Efros}, {and} \bibinfo{person}{Angjoo Kanazawa}.} \bibinfo{year}{2025}\natexlab{}.
\newblock \showarticletitle{Continuous 3D Perception Model with Persistent State}. In \bibinfo{booktitle}{\emph{IEEE Conf. Comput. Vis. Pattern Recog.}}
\newblock


\bibitem[Wang et~al\mbox{.}(2024a)]%
        {wang2024dust3r}
\bibfield{author}{\bibinfo{person}{Shuzhe Wang}, \bibinfo{person}{Vincent Leroy}, \bibinfo{person}{Yohann Cabon}, \bibinfo{person}{Boris Chidlovskii}, {and} \bibinfo{person}{Jerome Revaud}.} \bibinfo{year}{2024}\natexlab{a}.
\newblock \showarticletitle{Dust3r: Geometric 3d vision made easy}. In \bibinfo{booktitle}{\emph{IEEE Conf. Comput. Vis. Pattern Recog.}} \bibinfo{pages}{20697--20709}.
\newblock


\bibitem[Wizadwongsa et~al\mbox{.}(2021)]%
        {wizadwongsa2021nex}
\bibfield{author}{\bibinfo{person}{Suttisak Wizadwongsa}, \bibinfo{person}{Pakkapon Phongthawee}, \bibinfo{person}{Jiraphon Yenphraphai}, {and} \bibinfo{person}{Supasorn Suwajanakorn}.} \bibinfo{year}{2021}\natexlab{}.
\newblock \showarticletitle{Nex: Real-time view synthesis with neural basis expansion}. In \bibinfo{booktitle}{\emph{IEEE Conf. Comput. Vis. Pattern Recog.}} \bibinfo{pages}{8534--8543}.
\newblock


\bibitem[Xiong et~al\mbox{.}(2020)]%
        {on_layer_normalization}
\bibfield{author}{\bibinfo{person}{Ruibin Xiong}, \bibinfo{person}{Yunchang Yang}, \bibinfo{person}{Di He}, \bibinfo{person}{Kai Zheng}, \bibinfo{person}{Shuxin Zheng}, \bibinfo{person}{Chen Xing}, \bibinfo{person}{Huishuai Zhang}, \bibinfo{person}{Yanyan Lan}, \bibinfo{person}{Liwei Wang}, {and} \bibinfo{person}{Tie-Yan Liu}.} \bibinfo{year}{2020}\natexlab{}.
\newblock \showarticletitle{On layer normalization in the transformer architecture}. In \bibinfo{booktitle}{\emph{Int. Conf. Machine Learning}} \emph{(\bibinfo{series}{ICML'20})}. \bibinfo{publisher}{JMLR.org}, Article \bibinfo{articleno}{975}, \bibinfo{numpages}{10}~pages.
\newblock


\bibitem[Xu et~al\mbox{.}(2025)]%
        {xu2024depthsplat}
\bibfield{author}{\bibinfo{person}{Haofei Xu}, \bibinfo{person}{Songyou Peng}, \bibinfo{person}{Fangjinhua Wang}, \bibinfo{person}{Hermann Blum}, \bibinfo{person}{Daniel Barath}, \bibinfo{person}{Andreas Geiger}, {and} \bibinfo{person}{Marc Pollefeys}.} \bibinfo{year}{2025}\natexlab{}.
\newblock \showarticletitle{DepthSplat: Connecting Gaussian Splatting and Depth}. In \bibinfo{booktitle}{\emph{IEEE Conf. Comput. Vis. Pattern Recog.}}
\newblock


\bibitem[Xu et~al\mbox{.}(2023)]%
        {xu2023dmv3d}
\bibfield{author}{\bibinfo{person}{Yinghao Xu}, \bibinfo{person}{Hao Tan}, \bibinfo{person}{Fujun Luan}, \bibinfo{person}{Sai Bi}, \bibinfo{person}{Peng Wang}, \bibinfo{person}{Jiahao Li}, \bibinfo{person}{Zifan Shi}, \bibinfo{person}{Kalyan Sunkavalli}, \bibinfo{person}{Gordon Wetzstein}, \bibinfo{person}{Zexiang Xu}, {et~al\mbox{.}}} \bibinfo{year}{2023}\natexlab{}.
\newblock \showarticletitle{Dmv3d: Denoising multi-view diffusion using 3d large reconstruction model}.
\newblock \bibinfo{journal}{\emph{arXiv preprint arXiv:2311.09217}} (\bibinfo{year}{2023}).
\newblock


\bibitem[Yang et~al\mbox{.}(2025)]%
        {yang2025fast3r}
\bibfield{author}{\bibinfo{person}{Jianing Yang}, \bibinfo{person}{Alexander Sax}, \bibinfo{person}{Kevin~J Liang}, \bibinfo{person}{Mikael Henaff}, \bibinfo{person}{Hao Tang}, \bibinfo{person}{Ang Cao}, \bibinfo{person}{Joyce Chai}, \bibinfo{person}{Franziska Meier}, {and} \bibinfo{person}{Matt Feiszli}.} \bibinfo{year}{2025}\natexlab{}.
\newblock \showarticletitle{Fast3R: Towards 3D Reconstruction of 1000+ Images in One Forward Pass}. In \bibinfo{booktitle}{\emph{IEEE Conf. Comput. Vis. Pattern Recog.}}
\newblock


\bibitem[Yifan et~al\mbox{.}(2019)]%
        {yifan2019differentiable}
\bibfield{author}{\bibinfo{person}{Wang Yifan}, \bibinfo{person}{Felice Serena}, \bibinfo{person}{Shihao Wu}, \bibinfo{person}{Cengiz {\"O}ztireli}, {and} \bibinfo{person}{Olga Sorkine-Hornung}.} \bibinfo{year}{2019}\natexlab{}.
\newblock \showarticletitle{Differentiable surface splatting for point-based geometry processing}.
\newblock \bibinfo{journal}{\emph{ACM Trans. Graph.}} \bibinfo{volume}{38}, \bibinfo{number}{6} (\bibinfo{year}{2019}), \bibinfo{pages}{1--14}.
\newblock


\bibitem[Yu et~al\mbox{.}(2021a)]%
        {yu2021plenoctrees}
\bibfield{author}{\bibinfo{person}{Alex Yu}, \bibinfo{person}{Ruilong Li}, \bibinfo{person}{Matthew Tancik}, \bibinfo{person}{Hao Li}, \bibinfo{person}{Ren Ng}, {and} \bibinfo{person}{Angjoo Kanazawa}.} \bibinfo{year}{2021}\natexlab{a}.
\newblock \showarticletitle{Plenoctrees for real-time rendering of neural radiance fields}. In \bibinfo{booktitle}{\emph{Int. Conf. Comput. Vis.}} \bibinfo{pages}{5752--5761}.
\newblock


\bibitem[Yu et~al\mbox{.}(2021b)]%
        {yu2021pixelnerf}
\bibfield{author}{\bibinfo{person}{Alex Yu}, \bibinfo{person}{Vickie Ye}, \bibinfo{person}{Matthew Tancik}, {and} \bibinfo{person}{Angjoo Kanazawa}.} \bibinfo{year}{2021}\natexlab{b}.
\newblock \showarticletitle{pixelnerf: Neural radiance fields from one or few images}. In \bibinfo{booktitle}{\emph{IEEE Conf. Comput. Vis. Pattern Recog.}} \bibinfo{pages}{4578--4587}.
\newblock


\bibitem[Yu et~al\mbox{.}(2024)]%
        {yu2024gaussian}
\bibfield{author}{\bibinfo{person}{Zehao Yu}, \bibinfo{person}{Torsten Sattler}, {and} \bibinfo{person}{Andreas Geiger}.} \bibinfo{year}{2024}\natexlab{}.
\newblock \showarticletitle{Gaussian opacity fields: Efficient adaptive surface reconstruction in unbounded scenes}.
\newblock \bibinfo{journal}{\emph{ACM Trans. Graph.}} \bibinfo{volume}{43}, \bibinfo{number}{6} (\bibinfo{year}{2024}), \bibinfo{pages}{1--13}.
\newblock


\bibitem[Zhang et~al\mbox{.}(2024)]%
        {zhang2024gs}
\bibfield{author}{\bibinfo{person}{Kai Zhang}, \bibinfo{person}{Sai Bi}, \bibinfo{person}{Hao Tan}, \bibinfo{person}{Yuanbo Xiangli}, \bibinfo{person}{Nanxuan Zhao}, \bibinfo{person}{Kalyan Sunkavalli}, {and} \bibinfo{person}{Zexiang Xu}.} \bibinfo{year}{2024}\natexlab{}.
\newblock \showarticletitle{Gs-lrm: Large reconstruction model for 3d gaussian splatting}. In \bibinfo{booktitle}{\emph{Eur. Conf. Comput. Vis.}} Springer, \bibinfo{pages}{1--19}.
\newblock


\bibitem[Zhang et~al\mbox{.}(2018)]%
        {zhang2018unreasonable}
\bibfield{author}{\bibinfo{person}{Richard Zhang}, \bibinfo{person}{Phillip Isola}, \bibinfo{person}{Alexei~A Efros}, \bibinfo{person}{Eli Shechtman}, {and} \bibinfo{person}{Oliver Wang}.} \bibinfo{year}{2018}\natexlab{}.
\newblock \showarticletitle{The unreasonable effectiveness of deep features as a perceptual metric}. In \bibinfo{booktitle}{\emph{IEEE Conf. Comput. Vis. Pattern Recog.}} \bibinfo{pages}{586--595}.
\newblock


\bibitem[Zhou et~al\mbox{.}(2018)]%
        {zhou2018stereo}
\bibfield{author}{\bibinfo{person}{Tinghui Zhou}, \bibinfo{person}{Richard Tucker}, \bibinfo{person}{John Flynn}, \bibinfo{person}{Graham Fyffe}, {and} \bibinfo{person}{Noah Snavely}.} \bibinfo{year}{2018}\natexlab{}.
\newblock \showarticletitle{Stereo magnification: Learning view synthesis using multiplane images}.
\newblock \bibinfo{journal}{\emph{ACM Trans. Graph.}} (\bibinfo{year}{2018}).
\newblock


\bibitem[Ziwen et~al\mbox{.}(2025)]%
        {ziwen2024llrm}
\bibfield{author}{\bibinfo{person}{Chen Ziwen}, \bibinfo{person}{Hao Tan}, \bibinfo{person}{Kai Zhang}, \bibinfo{person}{Sai Bi}, \bibinfo{person}{Fujun Luan}, \bibinfo{person}{Yicong Hong}, \bibinfo{person}{Li Fuxin}, {and} \bibinfo{person}{Zexiang Xu}.} \bibinfo{year}{2025}\natexlab{}.
\newblock \showarticletitle{Long-LRM: Long-sequence Large Reconstruction Model for Wide-coverage Gaussian Splats}. In \bibinfo{booktitle}{\emph{Int. Conf. Comput. Vis.}}
\newblock


\end{thebibliography}
\begin{figure*}
  \includegraphics[width=0.95\linewidth]{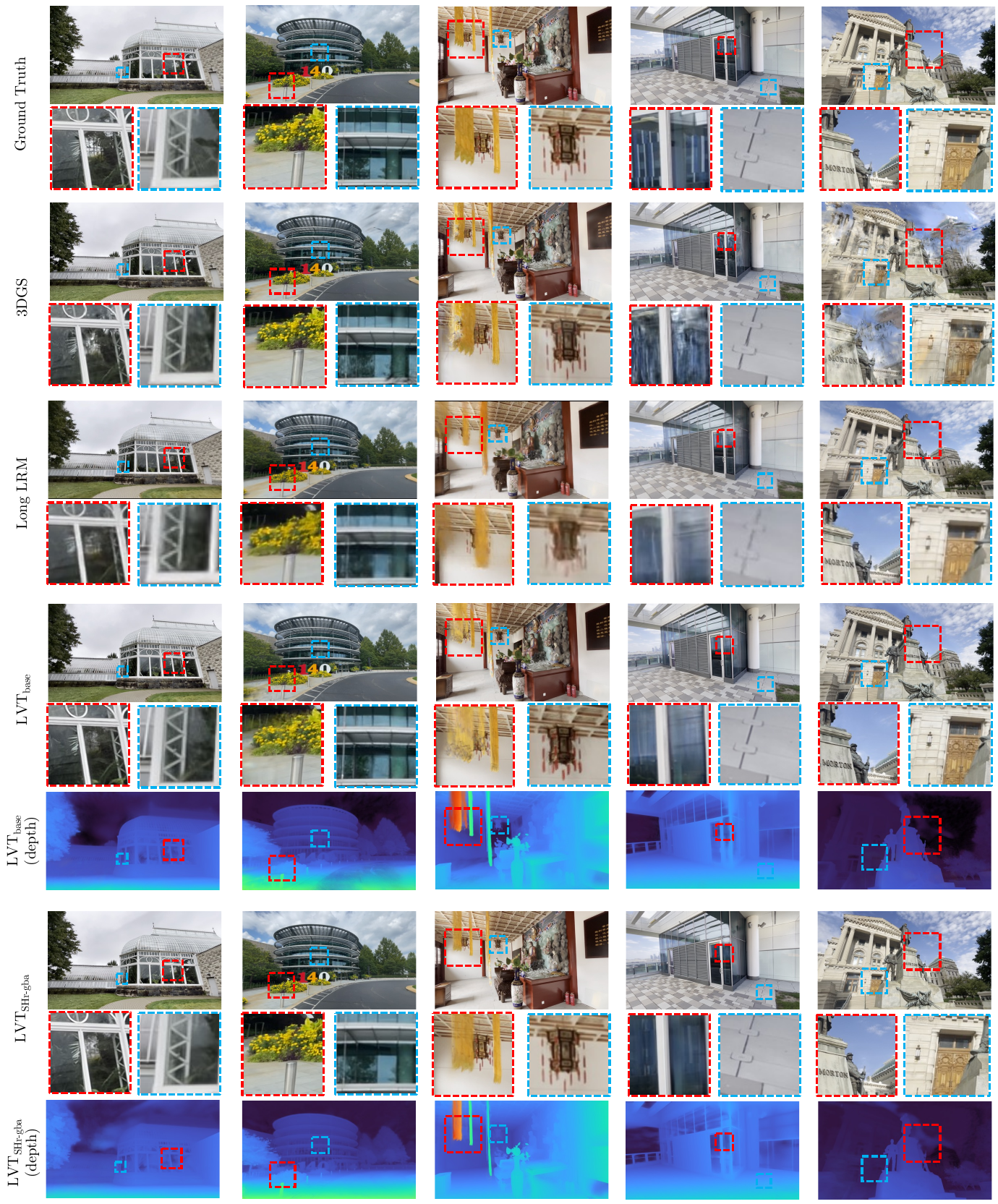}
  \caption{\textbf{Qualitative comparison of LVT and baselines on DL3DV.} We observe that 3DGS is prone to floater artifacts in reconstructed scenes. Long-LRM is less prone to floaters, but struggles to reconstruct thin structures (middle example) and specularities. LVT achieves high-fidelity reconstructions, and reconstructs specular reflections and thin structures particularly well when view-dependent opacity is introduced.}
  \label{fig:qual-dl3dv}
\end{figure*}

\clearpage
\setcounter{page}{1}

\begin{appendices}


\section{Additional Training Details}

\subsection{Dataset Mixing and Finetuning Strategy}\label{sec:supp-dataset-mixing}

\begin{figure}[h]
  \includegraphics[width=0.9\linewidth]{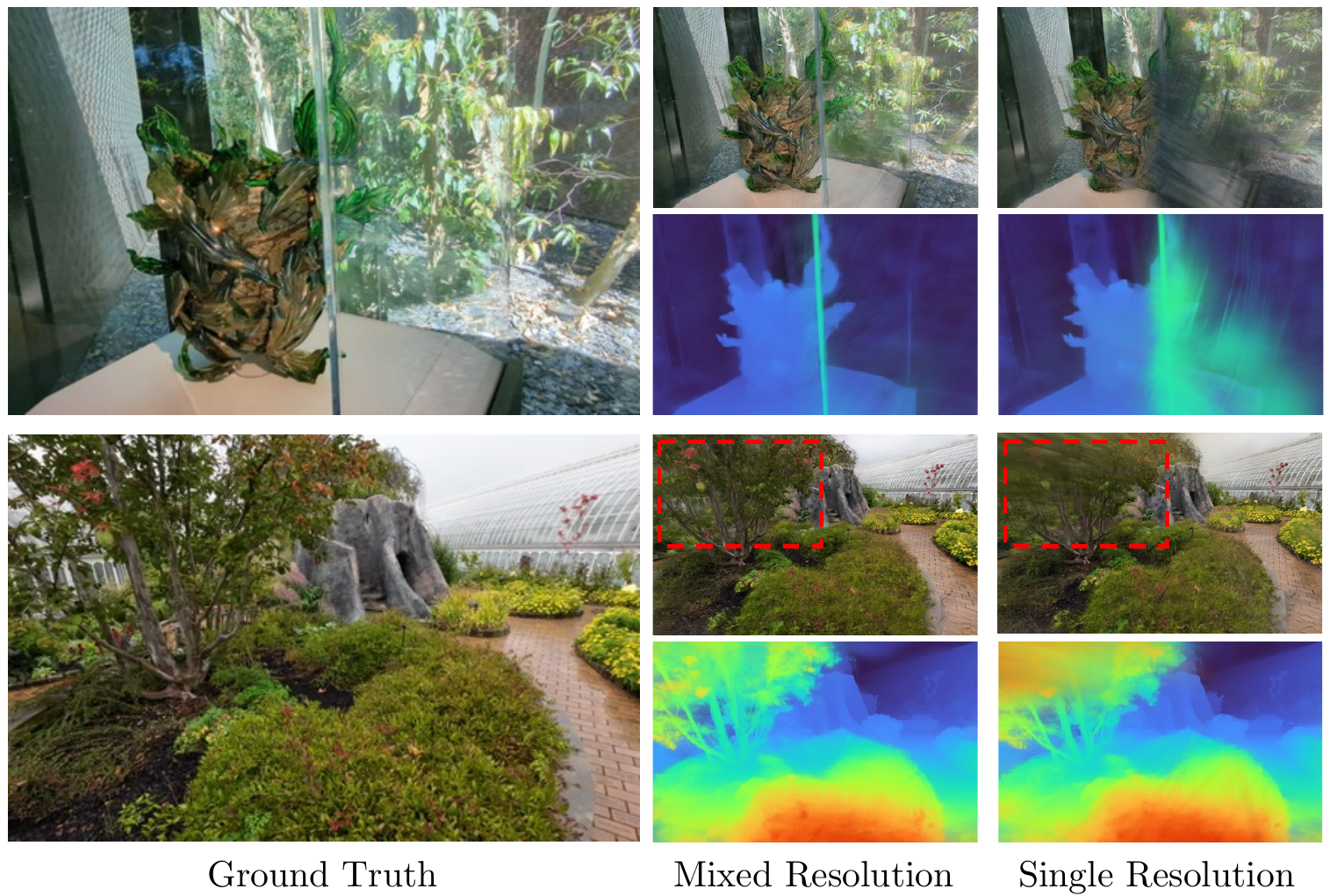}
  \vspace{-0.2cm}
  \caption{\textbf{Ablation for mixed-resolution training on DL3DV.} Our model trained with the mixed-resolution training strategy surpasses the model trained with a single input resolution. }
  \label{fig:mixres-ablation}
\end{figure}

We employ a dataset mixing strategy that randomly samples between longer sequences at lower resolution, and shorter sequences at higher resolution. This encourages the model to learn both high resolution details and scene priors on long scenes in a computationally feasible manner. We further detail our dataset mixing approach in Tab.~\ref{tab:supp_dataset}. We initially train our model to a maximum resolution of $480\times270$ and then finetune to $960\times540$. We show qualitative examples of the model output with and without this dataset mixing in Fig.~\ref{fig:mixres-ablation}. {To manage the memory footprint during finetuning, we cull 40\% and 20\% of splats with the smallest opacity for LVT$_{base}$ and LVT$_{SH-rgba}$ respectively. We found that LVT$_{base}$ is more prone to running out of memory during training, so we additionally finetune the model while randomly cropping target images to size of 672x378 and empirically reduce the LPIPS loss weight to 0.02 to balance the two reconstruction losses with this change in target image size.}

\begin{table}[h]
    \centering
    \resizebox{1.0\linewidth}{!}{
    \begin{tabular}{c |c c c c} 
         Model Resolution & Weight & Resolution &  \# Inputs & \# Targets \\
         \midrule
        \multirow{5}{*}{$480\times270$} 
        & 0.125 & 240x135 & 40 & 24\\
        & 0.125 & 240x135 & 44 & 24 \\
        & 0.125 & 240x135 & 48 & 24 \\
        & 0.125 & 240x135 & 52 & 24 \\
        & 0.5 & 480x270 & 16 & 8 \\
        \midrule
        \multirow{6}{*}{$960\times540$} 
        & 0.125 & 240x135 & 40 & 24\\
        & 0.125 & 240x135 & 44 & 24 \\
        & 0.125 & 240x135 & 48 & 16 \\
        & 0.125 & 240x135 & 52 & 16 \\
        & 0.25 & 480x270 & 24 & 2 \\
        & 0.25 & 960x540 & 16 & 1 \\
    \bottomrule
    \end{tabular}
    }
    \caption{\textbf{Dataset composition.} For mixed-resolution training, we train with longer sequences at lower resolution, and shorter sequences at higher resolution to learn both the necessary structure and details for scene reconstruction.}
    \label{tab:supp_dataset}
    \aftertable
\end{table}

\subsection{Model Training Time}
For DL3DV, pretraining takes 4 days at 480x270 resolution (with a batch size of 256, 1 per H100 GPU), followed by 2 days of fine-tuning at 960x540. We believe further optimizations can reduce this computational cost.

\subsection{Towards Higher Resolution with LVT-Flash}
Our model's computational cost scales linearly with the number of input images but quadratically with the number of pixels per image. This is because increasing the input resolution, while keeping the patch size constant, proportionally increases the number of tokens. For instance, doubling image width and height quadruples the tokens, leading to a 16-fold increase in transformer complexity, quickly becoming impractical at high resolutions.

\begin{figure}
  \includegraphics[width=\linewidth]{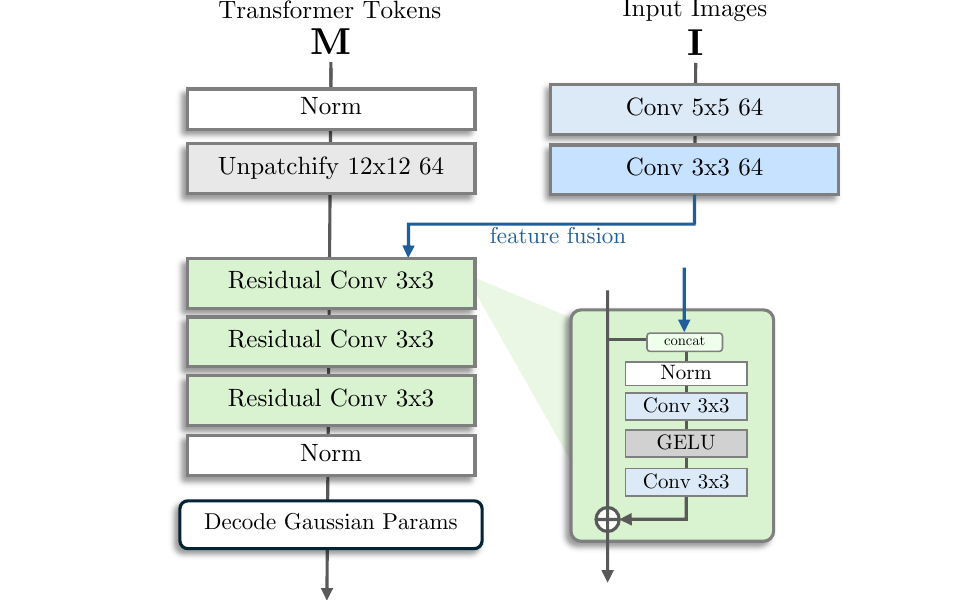}
  \caption{\textbf{Feature fusion for 2K outputs.} To produce higher resolution outputs, we use a larger patch size and then fuse in features from the input images to mitigate quality loss due to the increased patch size.}
  \label{fig:fusion}
\end{figure}

While larger patch sizes (e.g., using $12\times12$ instead of $8\times8$ patches) would reduce token count, this approach compromises quality as individual tokens lack the capacity to adequately encode larger input and output patches. To mitigate this quality loss, we introduce a post-processing step. After decoding transformer tokens to pixels, we fuse them with features extracted directly from the input images (Fig.~\ref{fig:fusion}). This allows high-resolution information from the input to enhance the decoded features. We call this faster, reduced quality model \emph{LVT-Flash}, and with it we can achieve 2k resolution in the transformer output.

\section{Additional Experiments}

\subsection{Model Inference Times}\label{sec:supp-inference-time}
\begin{figure}
  \includegraphics[width=\linewidth]{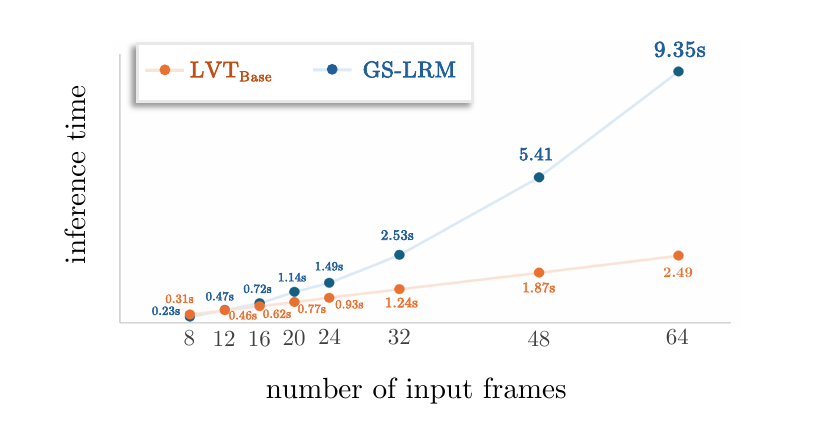}
  \caption{\textbf{Runtime comparison of local and global attention.} We evaluate all models on a single A100 40G GPU at 480x270 resolution. Our LVT model uses local attention and scales roughly linearly ($\mathcal{O}(N)$) with the number of input views N. We implement a model with full self-attention, following GS-LRM, which scales quadratically with the number of inputs ($\mathcal{O}(N^{2})$). This difference is also observed empirically in the model inference times.}
  \label{fig:time-comp-2}
\end{figure}

We compare the inference time of our model, with local attention windows, to a model with fully connected self attention, following the implementation of GS-LRM. Besides the tokenization and attention operations as described in Sec.~\ref{sec:method}, all other components and training of these two models are identical. We evaluate these two models, running in {16-bit} floating point, at 480x270 resolution and benchmark on a single A100 40G GPU.  In addition to finding that our LVT model is more robust to changes in sequence length at inference time, as reported in Tab.~\ref{tab:results_inputs}, the inference time our model also scales linearly with sequence length, whereas the self attention model exhibits dramatically longer inference time as sequence length increases (Fig.~\ref{fig:time-comp-2}).

\begin{figure}
  \includegraphics[width=0.93\linewidth]{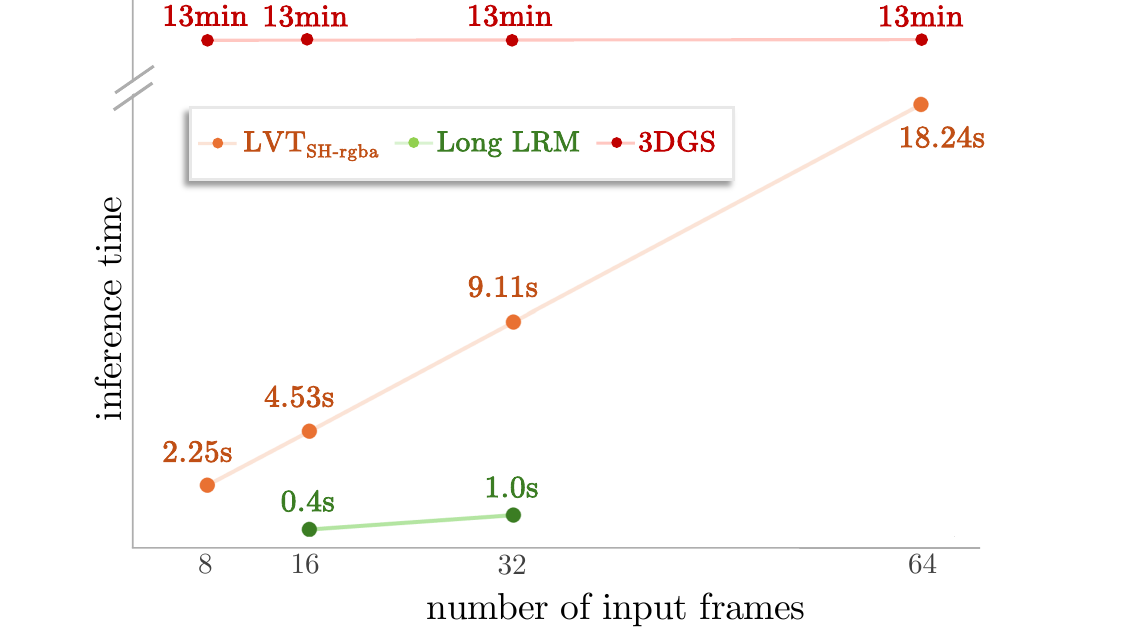}
  \caption{\textbf{Inference time comparison to baselines.} We evaluate all models on a single A100 40G GPU at 960x540 resolution. LVT$_{SH-rgba}$ scales roughly linearly ($\mathcal{O}(N)$) with the number of input views N. Long‑LRM, despite a lower constant factor from its Mamba–attention hybrid, still includes global self‑attention operation. The optimization time of 3DGS depends on the number of optimization steps.}
  \label{fig:time-comp}
\end{figure}

We also benchmark our full model trained at 540x960 resolution in a similar fashion. 3DGS requires optimization per scene, and its runtime is dominated by the number of optimization steps rather than the number of inputs. While our model is not as fast as the times reported by Long-LRM, we note their model is a hybrid of Mamba and transformer blocks, with the transformer block still operating at quadratic complexity. In contrast, our model uses transformer blocks only with local attention, and for model blocks with memory-intensive intermediate states, operations can be computed sequentially on a per-image basis rather than processing all images at once. This incremental approach effectively reduces memory consumption by trading off some computational speed. Again, we observe a linear trend between input sequence length and runtime for our model, significantly father than optimization-based 3DGS (Fig. ~\ref{fig:time-comp}).


\subsection{RealEstate-10k}\label{sec:supp-re10k}

We also train a version of our model on RealEstate-10K~\cite{zhou2018stereo} to benchmark our model against GS-LRM~\cite{zhang2024gs}. Following prior work, we train at $256\times256$ resolution. We use loss weights $\lambda=0.25$ and $\alpha=0$ and the same learning rate schedule as our DL3DV model.



\begin{figure*}[h]
  \includegraphics[width=0.98\linewidth]{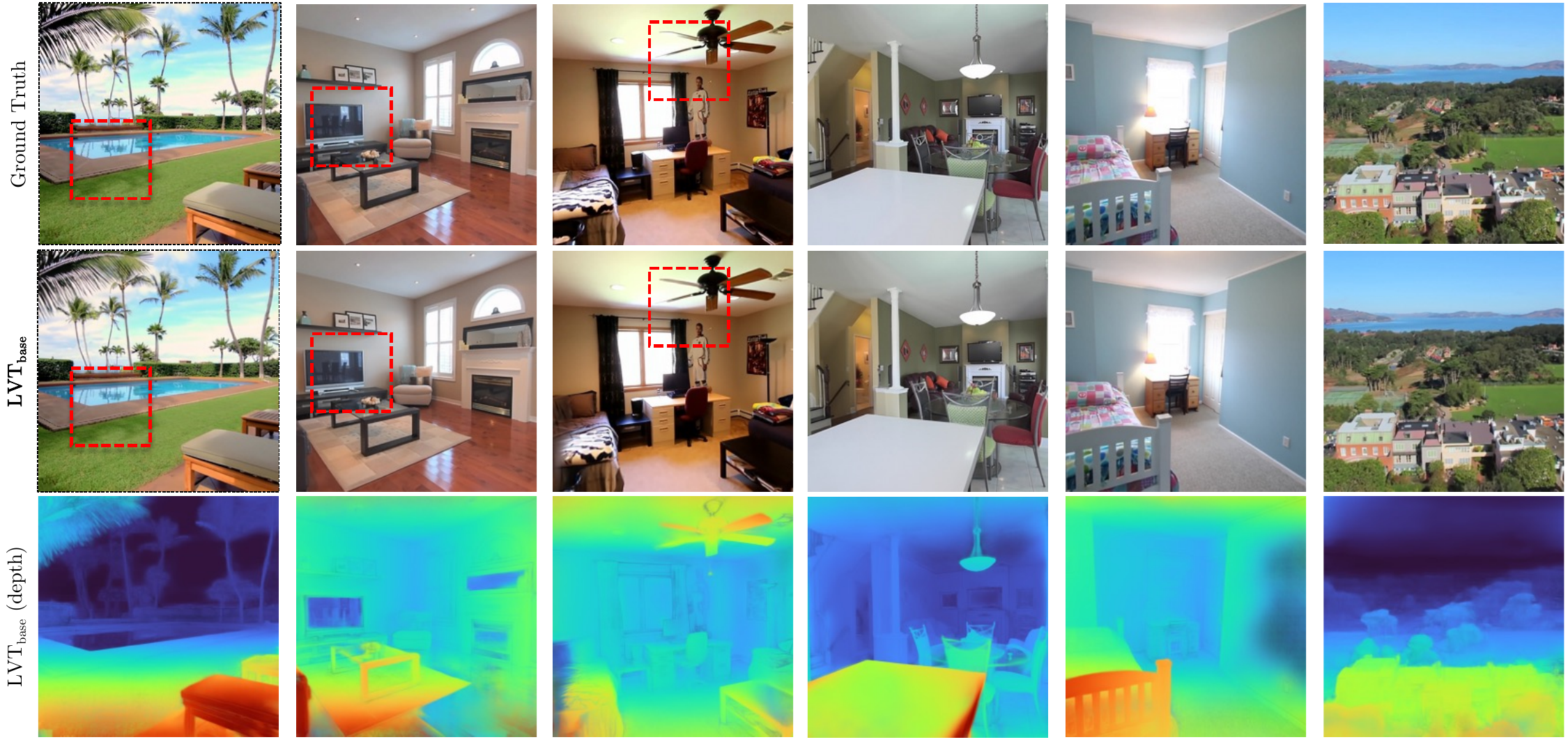}
  \vspace{-0.2cm}
  \caption{\textbf{Qualitative results on RealEstate10K.} We show qualitative results of our LVT model trained on 2 input frames of RealEstate10K dataset at $256\times256$ resolution.}
  \label{fig:qual-re10k}
\end{figure*}

\myparagraph{Two-view model.} We train a version of our LVT$_{base}$ model on two input views. In this case the neighbor set only contains two items per view, itself and the other input view. While this is not the intended use-case of our model, we conduct this experiment to ensure that our conditioning and attention modules do not degrade performance compared to the standard transformer architecture, as used in GS-LRM. We use the standard training/test split (comprising $\sim$ 90\%/10\% clips respectively) and input and target indices as adopted in Pixel-Splat~\cite{charatan2024pixelsplat}. We compare to previous generalizable methods~\cite{yu2021pixelnerf, suhail2022generalizable, du2023learning,zhang2024gs, ziwen2024llrm, charatan2024pixelsplat} in Tab.~\ref{tab:results_re10k}. Our model obtains competitive performance with GS-LRM and Long-LRM on this simplified setting, ensuring that our changes to the model do not hinder the transformer. We show qualitative results from our model in Figure \ref{fig:qual-re10k}.



\begin{table}[ht]
    \centering
    \begin{tabular}{l|c c c} 
         Model & \textit{PSNR} $\uparrow$ & \textit{SSIM}  $\uparrow$ & \textit{LPIPS}  $\downarrow$ \\
         \midrule
         pixelNeRF  & 20.43 & 0.589 & 0.550 \\
         GPNR       & 24.11 & 0.793 & 0.255 \\
         Du et al.  & 24.78 & 0.820 & 0.213 \\
         pixelSplat & 25.89 & 0.858 & 0.142 \\
         GS-LRM     & 28.10 & 0.892 & 0.114 \\
         Long-LRM (no token merge) & {28.54} & {0.895} & {0.109} \\
         LVT$_{base}$ (Ours)  &\textbf{ 28.75} & \textbf{0.915} & \textbf{0.072} \\
    \bottomrule
    \end{tabular}
    \caption{\textbf{RealEstate10K evaluations.} We compare LVT against baseline methods at a resolution of $256\times256$. We use the evaluation split from pixelSplat~\cite{charatan2024pixelsplat}, which takes two images as input.}
    \label{tab:results_re10k}
    \aftertable
\end{table}

\begin{figure}
  \includegraphics[width=\linewidth]{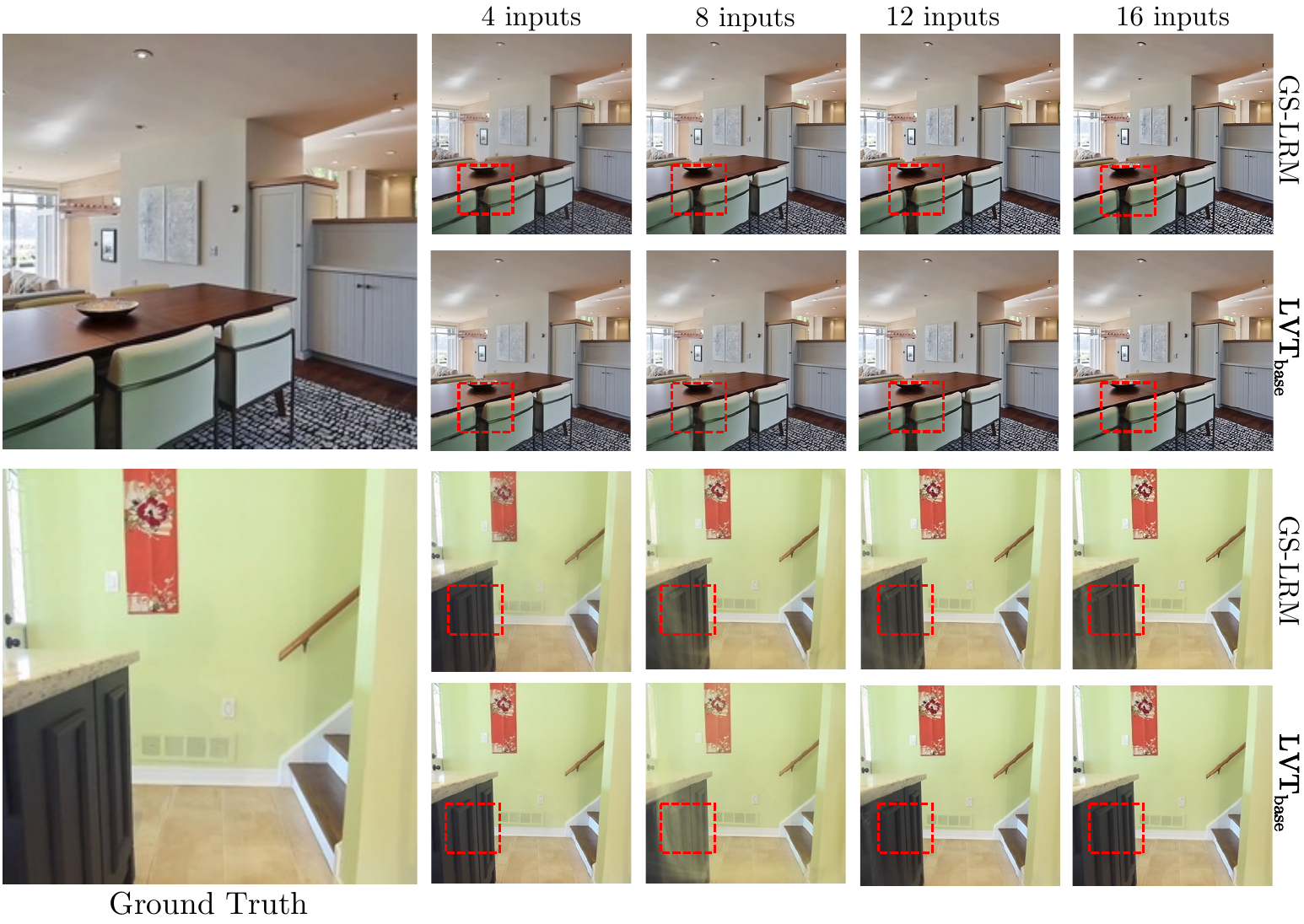}
  \vspace{-0.2cm}
  \caption{\textbf{Ablation for generalization to variable sequence lengths on RealEstate10K.} We train both models on 8 input frames and test. LVT effectively consolidates information from more input views.}
  \label{fig:re10k-seq-len-abl}
\end{figure}

\myparagraph{Sequence model.} Additionally, to demonstrate the generalization ability of LVT to varying sequence lengths, we train our LVT$_{base}$ model with a window size of three neighbors and a full self-attention model. The full self-attention model is based on GS-LRM, which concatenates world-space Pl\"ucker rays to the input images and uses standard transformer blocks. All other components of the models are identical, and we train both using 8 input views. As reported in Table~\ref{tab:results_re10k_generalization} and Figure~\ref{fig:re10k-seq-len-abl}, we evaluate both models on scenes by feeding 4, 8, 12 and 16 input views. Notably, the performance difference between the two models widens as the discrepancy between the number of training and evaluation input views increases, showing that our model generalizes better to varied sequence lengths. 

\begin{table}
    \centering
    \begin{tabular}{l|c|c c c} 
         Model & \# of inputs & \textit{PSNR} $\uparrow$ & \textit{SSIM}  $\uparrow$ & \textit{LPIPS}  $\downarrow$ \\
         \midrule
         \multirow{4}{*}{Full SA} & 4 & 28.02 & 0.929 & 0.064 \\
         & 8 & 27.81 & 0.922 & 0.079 \\
         & 12 & 28.15 & 0.932 & 0.071 \\
         & 16 & 26.72 & 0.922 & 0.080 \\
         \midrule
          & 4 & \textbf{31.67} & \textbf{0.951} & \textbf{0.046} \\
         LVT$_{base}$ & 8 & 27.80 & 0.921 & 0.078 \\
         (Ours) & 12 & 29.00 & 0.932 & 0.070 \\
         & 16 & 29.46 & 0.936 & 0.066 \\
    \bottomrule
    \end{tabular}
    \caption{\textbf{RealEstate10K evaluations with variable sequence lengths.} We train a full self-attention model, following GS-LRM and LVT on 8 input views, but evaluate on varying number to inputs to assess their generalization ability for changing input sequence lengths. We use a resolution of $256\times 256$ throughout.}
    \label{tab:results_re10k_generalization}
    \aftertable
\end{table}

\subsection{Additional Ablations}\label{sec:supp-ablation}

We investigate additional design decisions of our LVT architecture via ablation experiments, computed with batch size 64 at 480x270 resolution.

\begin{table}
    \centering
    \begin{tabular}{c|c|c c c} 
         Window Size & Dilation & \textit{PSNR} $\uparrow$ & \textit{SSIM}  $\uparrow$ & \textit{LPIPS}  $\downarrow$ \\ 
         \midrule
         3 & 1 & 21.75 & 0.765 & 0.205 \\
         5 & 1 & 22.02 & 0.774 & 0.196 \\
         5 & 2 & 21.26 & 0.746 & 0.221 \\
         7 & 1 & 21.91 & 0.772 & 0.198 \\
    \bottomrule
    \end{tabular}
    \caption{\textbf{Ablation for neighborhood window sizes and dilations in the local view transformer.} We evaluate various window sizes with and without dilation in neighbor selection, where a dilation of 1 corresponds to no dilation/selecting directly adjacent neighbors. A neighborhood of window size 5 with no dilation is the best-performing variant.}
    \label{tab:abl-window}
\end{table}

\begin{table}
    \centering
    \begin{tabular}{l|c|c c c} 
         Ablation & Relative SH & \textit{PSNR} $\uparrow$ & \textit{SSIM}  $\uparrow$ & \textit{LPIPS}  $\downarrow$ \\ 
         \midrule
         LVT$_{base}$ & - & 22.02 & 0.774 & 0.196 \\
         + SH color & \xmark & 20.08 & 0.707 & 0.243 \\
          & \cmark & 22.48 & 0.795 & 0.181 \\
         + SH opacity & \xmark & 19.56 & 0.676 & 0.291 \\
          & \cmark & 23.89 & 0.828 & 0.152 \\
    \bottomrule
    \end{tabular}
    \caption{\textbf{Ablation for view-dependent splat color and opacity.}  Incorporating view dependence into both color and opacity consistently boosts reconstruction quality compared to the baseline. The SH need to be computed relative to the input cameras (instead of world space, which can be arbitrarily defined).}
    \label{tab:abl-sh}
\end{table}

\myparagraph{Neighborhood size and dilation.} On DL3DV, for 32 input frames at 480x270 resolution, our LVT model uses a neighborhood size of five neighbors. Here, we investigate the impact of this parameter, as well as the dilation factor.  A dilation of 1 indicates that only the N closest neighbors are attended to, whereas a dilation of 2 takes every other of the 2N closest neighbors. Results are shown in Tab.~\ref{tab:abl-window}.
We find that reducing the window size to 3 or increasing it to 7 with no dilation leads to a marginal decrease. Introducing a dilation factor of 2 with a window size of 5 further degrades performance, implying that considering non-adjacent neighbors within the local window is detrimental to the reconstruction quality. Moreover, with no dilation the inference times are 0.95s, 1.27s, and 1.62s for neighborhood window sizes 3, 5 and 7, respectively. As such, we default to a neighborhood window size of 5 without dilation for our main model.

\begin{figure*}
  \includegraphics[width=0.9\linewidth]{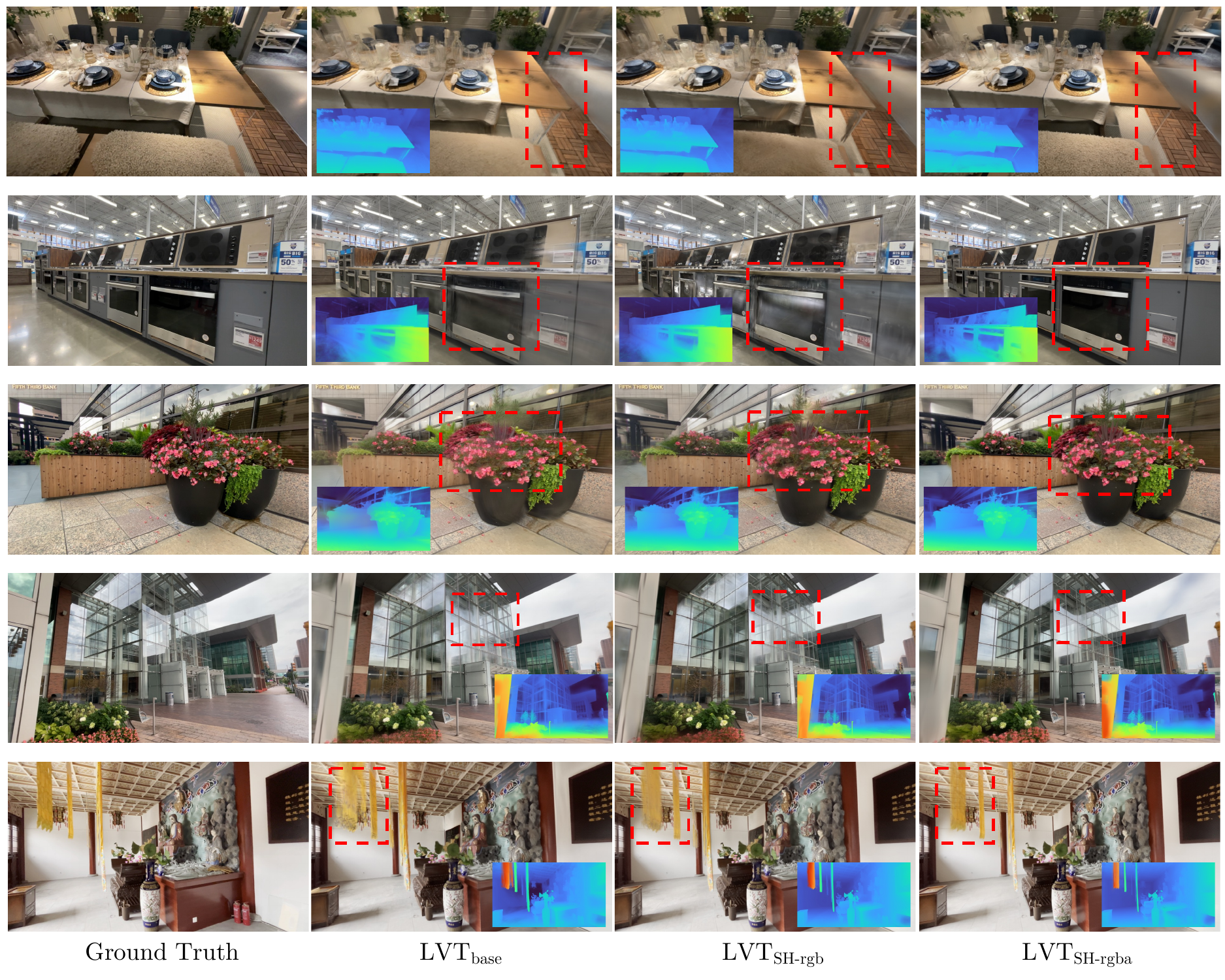}
 \vspace{-0.2cm}
  \caption{\textbf{Impact of view dependence.} We show qualitative examples of our model adding spherical harmonics on color, and on both color and opacity. LVT$_{SH-rgba}$ attains the best metrics and is better able to represent complex structures and lighting effects.}
  \label{fig:view-dep-abl}
\end{figure*}

\myparagraph{View-dependent splat parameters.} We present results on incorporating view dependence into the color and opacity predictions of our model in Tab.~\ref{tab:abl-sh} and Fig.~\ref{fig:view-dep-abl}.
We find that maintaining transformation invariance by predicting spherical harmonics coefficients relative to each input camera's local coordinate system (Relative SH in Tab.~\ref{tab:abl-sh}) is critical. Without it, the model cannot generalize to different sequence lengths at inference time.

\begin{figure}
  \includegraphics[width=\linewidth]{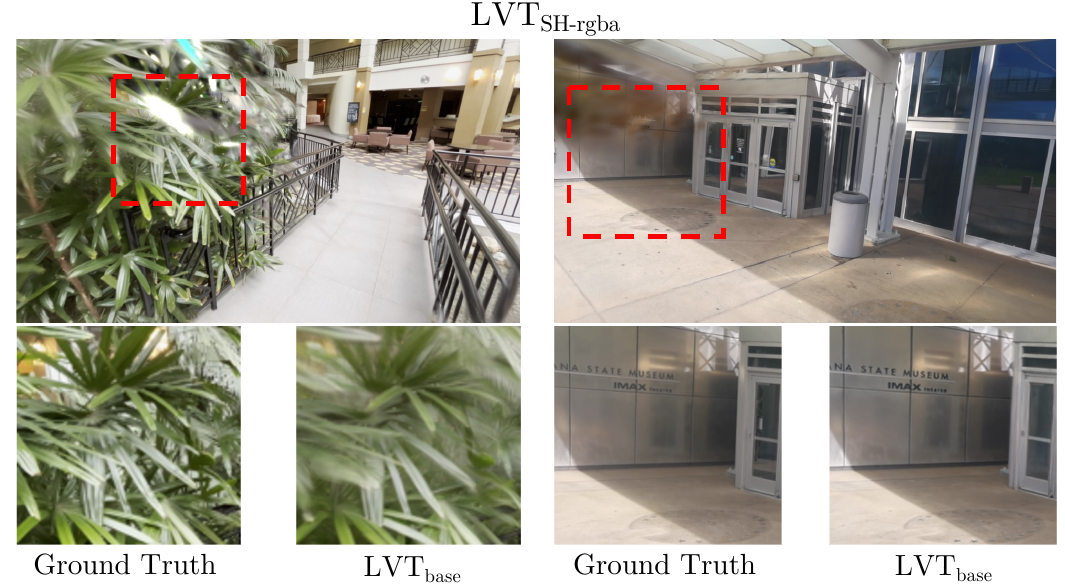}
  \vspace{-0.2cm}
  \caption{\textbf{Limitations of view-dependent opacity.} Using the view-dependent opacity occasionally produces `popping' artifacts in regions with extreme (highly dense or smooth) textures.}
  \label{fig:failure-cases}
\end{figure}

Adding spherical harmonics on color improves over our LVT$_{base}$ model, and adding view-dependent opacity offers further improvements. However, we occasionally observe some failure modes in which the view-dependent opacity can cause `popping' artifacts, as shown in Fig.~\ref{fig:failure-cases}.

\myparagraph{Input Camera Selection.} For our experiments, we default to selecting every eighth image in the full image sequence as input images to the model, after all target images have been removed. This varies the distance of the selected input images to the target cameras, without requiring more advanced heuristics that maintain a fixed sequence length. In Table~\ref{tab:abl-inputs}, we test our 960x540 resolution models while varying the input selection stride at inference time, taking every sixth, eighth, or tenth image as input. Our models are relatively robust to this selection, and we choose every eighth as a good middle ground.

\begin{table}
    \centering
    \begin{tabular}{l|c|c c c} 
         Model & Input Stride & \textit{PSNR} $\uparrow$ & \textit{SSIM}  $\uparrow$ & \textit{LPIPS}  $\downarrow$ \\ 
         \midrule
        \multirow{3}{*}{ LVT$_{base}$} & 6 & 23.18 & 0.810 & 0.221 \\ 
          & 8 & 23.65 & 0.813 & 0.215 \\
          & 10 & 23.67 & 0.807 & 0.218 \\
        \midrule
        \multirow{3}{*}{ LVT$_{SH-rgba}$} 
         & 6 & 27.95 & 0.891 & 0.127 \\
         & 8 & 27.64 & 0.883 & 0.133 \\
         & 10 & 27.13 & 0.873 & 0.141 \\
    \bottomrule
    \end{tabular}
    \caption{\textbf{Varying input image selection.}  We vary the stride of the input images, selecting every sixth, eighth, or tenth image in the sequence as model inputs after all target images have been removed. Our models are relatively robust to the input selection, which in turn changes the overall total number of input images per scene.}
    \label{tab:abl-inputs}
\end{table}

\begin{figure*}
  \includegraphics[width=\linewidth]{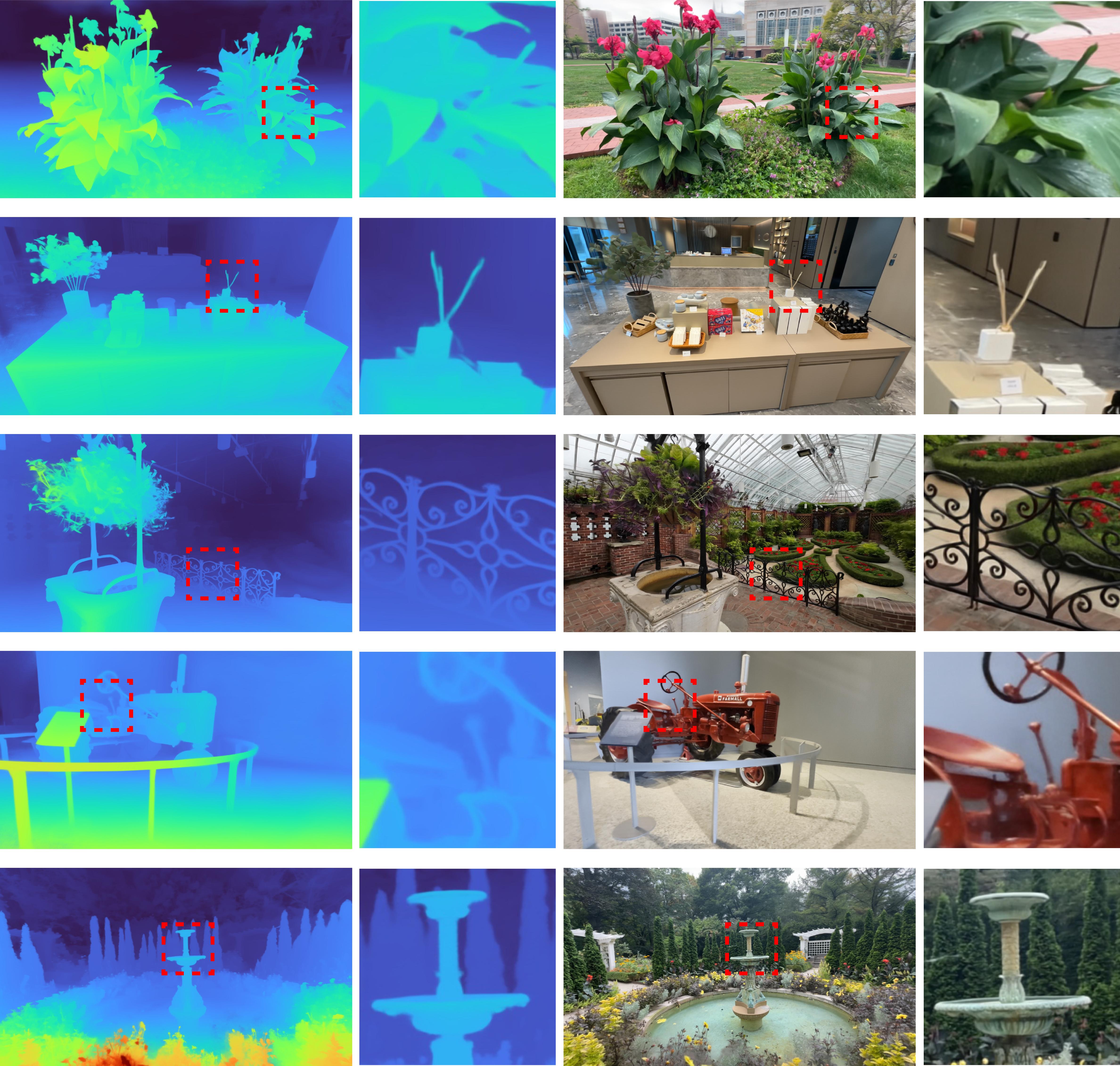}
  \vspace{-0.2cm}
  \caption{\textbf{\textit{LVT-Flash} - DL3DV-140 at 2K resolution. }LVT-Flash enables efficient processing of high-resolution images by employing larger patch sizes, which reduces the number of tokens to be processed. To counteract potential quality degradation from these larger patches, our proposed feature fusion module integrates high-resolution details from the input image directly with the decoded transformer tokens, thereby preserving output detail and enabling high-resolution results.}
  \label{fig:lvtflash}
\end{figure*}

\myparagraph{Higher resolution with LVT-Flash.} In Fig.~\ref{fig:lvtflash} we show outputs from our LVT-Flash model. Qualitatively we find that using a feature fusion module can recover some of the quality loss incurred from increasing the patch size and remains efficient enough to render at 2K resolution compared to using a smaller patch size.




\end{appendices}

\end{document}